\documentclass[authoryear,preprint,review,12pt,authoryear, 3p]{elsarticle}

\usepackage{amssymb}
\usepackage{amsthm}

\usepackage{hyperref}
\usepackage{xcolor}
\usepackage{color}
\usepackage{natbib}
\usepackage{dcolumn}
\usepackage{bm}
\usepackage{booktabs}
\usepackage{multirow}
\usepackage{framed}
\usepackage{siunitx}
\usepackage{nomencl}
\usepackage{ifthen}
\usepackage{etoolbox}
\usepackage{titlesec}
\usepackage{url}
\usepackage{algpseudocode}
\usepackage{algorithm}
\usepackage{natbib}
\usepackage[english]{babel}
\usepackage{amsmath}
\usepackage{amsfonts}
\usepackage{upgreek}
\usepackage{amssymb}
\usepackage{graphicx}
\usepackage{caption}
\usepackage{subcaption}
\usepackage{float}
\usepackage{mathtools}

\def\ps@pprintTitle{%
\let\@oddhead\@empty
\let\@evenhead\@empty
\def\@oddfoot{\footnotesize\itshape
 \ifx\@journal\@empty Elsevier
\else\@journal\fi\hfill\today}%
\let\@evenfoot\@oddfoot}

\begin{document}

\begin{frontmatter}



\title{An Analysis of Multi-Agent Reinforcement Learning for Decentralized Inventory Control Systems}


\affiliation[inst1]{organization={Sargent Centre for Process Systems Engineering},
            addressline={Department of Chemical Engineering, Imperial College London}, 
            city={London},
            postcode={SW7 2AZ}, 
            country={United Kingdom}}
            
\affiliation[inst2]{organization={Centre for Process Integration},
            addressline={Department of Chemical Engineering, The University of Manchester, Manchester}, 
            city={Manchester},
            postcode={M13 9PL}, 
            country={United Kingdom}}

\author[inst1]{Marwan Mousa}
\author[inst1]{Damien van de Berg}
\author[inst1]{Niki Kotecha}
\author[inst1]{Ehecatl Antonio del Rio-Chanona}
\author[inst2]{Max Mowbray}

\begin{abstract}
    Most solutions to the inventory management problem assume a centralization of information that is incompatible with  organisational constraints in real supply chain networks. The inventory management problem is a well-known planning problem in operations research, concerned with finding the optimal re-order policy for nodes in a supply chain. While many centralized solutions to the problem exist, they are not applicable to real-world supply chains made up of independent entities. The problem can however be naturally decomposed into sub-problems, each associated with an independent entity, turning it into a multi-agent system. Therefore, a decentralized data-driven solution to inventory management problems using multi-agent reinforcement learning is proposed where each entity is controlled by an agent. Three multi-agent variations of the proximal policy optimization algorithm are investigated through simulations of different supply chain networks and levels of uncertainty. The centralized training decentralized execution framework is deployed, which relies on offline centralization during simulation-based policy identification, but enables decentralization when the policies are deployed online to the real system. Results show that using multi-agent proximal policy optimization with a centralized critic leads to performance very close to that of a centralized data-driven solution and outperforms a distributed model-based solution in most cases while respecting the information constraints of the system.
\end{abstract}




\end{frontmatter}



\section{Introduction}
\label{sec: Introduction}

\subsection{An overview of supply chain management and uncertainty}

Production planning, inventory control and transportation form key decision functions within industrial production systems. These functions identify decisions that define production targets, as well as maintain and transport material inventory, across multiple distributed production facilities (i.e. echelons) to transform raw materials into marketable products. These decisions may be continuous or discrete depending on the decision function, and as a result, these problems are formulated either as either linear programming (LP) or mixed-integer linear programming (MILP). Indeed, exact optimization approaches underpin the current state-of-the-art for industrial scale problems \citep{grossmann2016recent}. 

However, these decision problems are uncertain, with the associated random variables either exogenous or endogenous in nature. In multi-echelon inventory control problems, common respective examples include customer demand at the retailer and production lead times \citep{song2017optimal}. To mitigate phenomena such as the bull-whip effect \citep{Bullwhip} in inventory control problems, decision-making should be coordinated across the supply chain and account for uncertainty. A number of stochastic \citep{hamdan2019two}, robust  \citep{aharon2009robust} and distributionally-robust \citep{hashemi2023integrated} mathematical programming formulations have been proposed to account for distributional, set-based, and ambiguity set descriptions of uncertain parameters, respectively. Despite providing principled frameworks to account for uncertainty, stochastic and distributionally-robust approaches face issues related to online tractability, with robust formulations known to introduce conservatism, when applied within receding-horizon frameworks.

\subsection{Reinforcement Learning and multi-echelon inventory control}

More recently, there has been interest in the development of Reinforcement Learning (RL) solutions to multi-echelon inventory control problems \citep{boute2022deep}. Closely connected to dynamic programming (DP), RL has been demonstrated as a general solution method for identifying approximately optimal parametric policy functions for stochastic decision processes. This is primarily because parametric RL policies aim to satisfy the Bellman optimality equation \citep{powell2007approximate, bhandari2019global} with optimal policy parameters identified through iterative model-free updates (e.g. via the policy gradient theorem) \citep{levine2020offline}. This process can be thought of as a noisy search process. In principle, it could be conducted online \citep{lawrence2022deep}, but for the purpose of safety and sample efficiency preference is for identification to utilize offline simulation of an approximate model typically with distributional descriptions of uncertain parameters \citep{mowbray2022industrial}. This enables the cheap generation of near-optimal online management decisions by the inference process of the policy function approximation. This is a major benefit relative to mathematical programming approaches that require resolving an optimization problem recursively in a receding or shrinking horizon framework. This requirement often means that, for example, stochastic mathematical programming models have to make significant approximation to uncertain parameters described by large or continuous support in order satisfy time constraints imposed on the identification of a decision.

Based on the promise of RL, a number of preliminary studies have been conducted. For example, \cite{hubbs2020orgym} demonstrated the application of the proximal policy optimization (PPO) algorithm on a multi-echelon inventory control problem subject to demand uncertainty and integer constraints on reorder decisions. The algorithm was demonstrated to outperform a MILP strategy with nominal demand data. \cite{wu2023distributional} provided a derivative-free optimization approach to RL together with a flexible, risk-sensitive formulation. The method was benchmarked to the work provided in \cite{hubbs2020orgym} and demonstrated an 11\% improvement in expected performance for the same computational budget. \cite{ORgymextension} extended this analysis for continuous reorder decisions and benchmarked against nominal and multi-stage stochastic linear programming formulations under the assumption of demand uncertainty at the retailer. 

In general, these investigations demonstrated that the RL methods were competitive with but outperformed by benchmark mathematical programming formulations. However, although these works provide an excellent investigation of RL, they consider supply chain instances with sequential structure, relatively few production echelons, and make two important assumptions. Firstly, production processes and transportation times are jointly approximated via a deterministic safety lead time. This is to overcome the computational barriers of integrating the description of planning, production, and transportation into one model and is a common assumption in constructing supply chain management policies \citep{lejarza2022feedback}. However, the rationale for describing lead time as a random variable has long been documented. For example, \cite{nevison1984dynamic} presented a DP approach to single-product, single-echelon inventory control under exogenous production lead times and deterministic demand. It should be noted that the application of DP necessitates a relatively simple problem instance with small control and state sets. \cite{song2017optimal} characterize the optimal reorder policy for a single product, single-echelon dual-sourcing problem, but with endogenous lead time uncertainty. 

For larger industrial problems, similar reasoning processes to that provided in \cite{song2017optimal} become very difficult and reliant on advanced computational tools. 
\cite{thevenin2022robust} proposed a robust optimization approach to identify a multi-period single-echelon inventory reorder policy subject to lead time uncertainty and multi-sourcing (i.e. with at least two or more potential suppliers). \cite{liu2021two} proposed a two-stage, distributionally robust model to handle uncertainty in transportation times for large-scale maritime inventory routing problems. The authors demonstrate that a tailored multi-cut bender's decomposition algorithm can exploit structure in the resultant model to reduce the computation required to identify a solution. \cite{franco2020optimization} proposed a stochastic formulation considering lead time uncertainty for inventory control of multiple pharmaceutical products within a hospital. The policy derived from the methodology provides a performance improvement of 15\% to that currently implemented by the hospital subject to the case study. However, as reviewed in \cite{ben2022supply} and highlighted in \cite{franco2020optimization}, it is rare that lead time uncertainty is accounted for in the literature. To the author's knowledge, only two works detailed by \cite{gijsbrechts2022can} and \cite{madeka2022deep} consider the impact of stochastic lead times on parametric inventory control policy approximations. However, analysis in both is specific to single-echelon inventory control problems.

The second assumption of previous works examining the use of RL is that information can be centralized for the purposes of decision-making in real-time to enable closed-loop decision-making. As the scale of the system increases, this assumption will become more difficult to satisfy in practice, primarily because the nature of production is inherently distributed \citep{andersson2000decentralized, ghasemi2022coordination}, but also due to information sharing constraints often observed in multi-stage production systems \citep{sahin2002flow}. This has led to an interest in the development of decentralized decision-making frameworks.

\subsection{Decentralized decision-making and multi-echelon inventory control}

\begin{figure}
    \centering
    \includegraphics[width=0.95\linewidth]{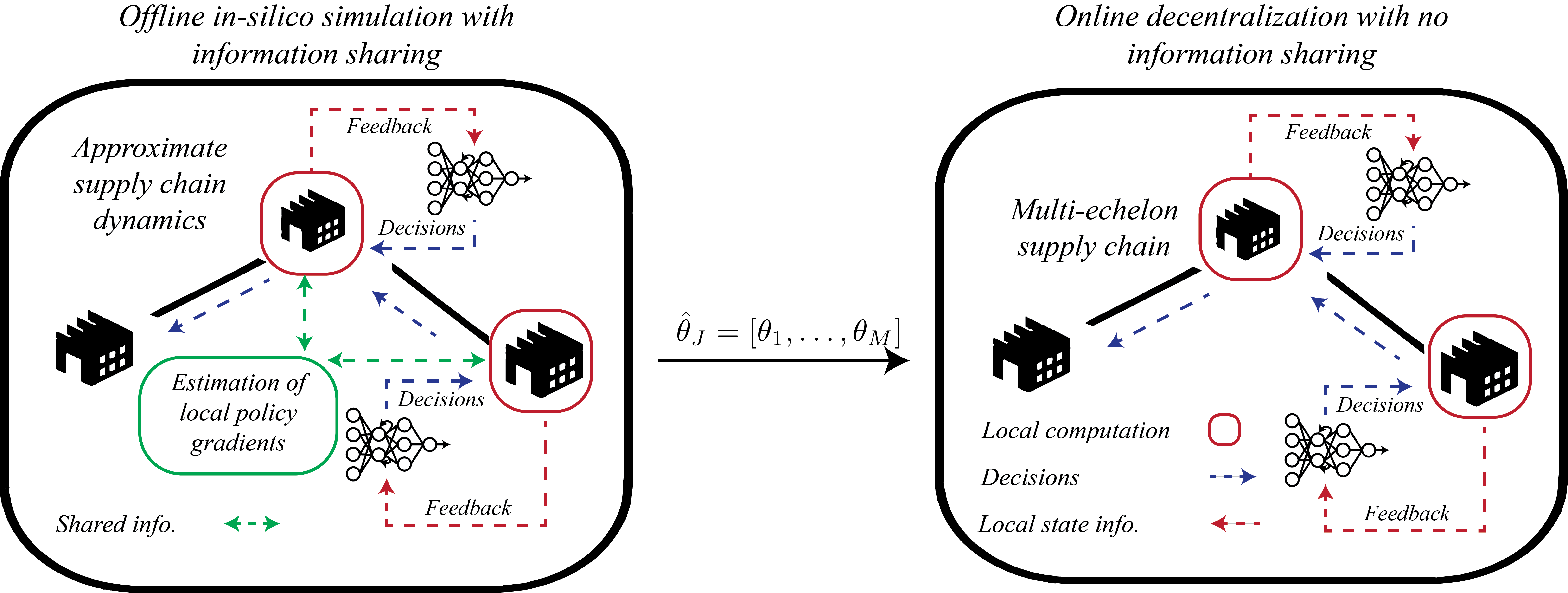}
    \caption{A framework for coordinating multi-agent reinforcement learning policies offline, enabling complete decentralization of decision-making online, through transfer of policy parameters, $\hat{\theta}_J$}
    \label{fig:cdte}
\end{figure}

Two class of methods that could reduce dependence on the latter assumption is distributed optimization and multi-agent RL (MARL). The use of distributed optimization approaches for inventory control has been well described elsewhere \citep{DunbarMPCBDG, CoopDMPCSC}. A major comparative benefit of MARL approaches is that they inherit flexibility in the treatment of sources of uncertainty associated with single agent RL and other simulation-based approaches, but enable the decentralization of collaborative decision-making in deployment, without the requirements for coordinated computation online. Multiple agents hold independent parametric policies, which are each conditioned on a local, partial state observation to provide a decentralized joint policy. This means the state information does not need to be shared between production echelons online. As a result, MARL provides an opportunity to coordinate decision-making in a decentralized manner, whilst respecting information-sharing constraints online. Instead, multi-agent coordination is provided during policy identification, which is conducted offline. This is highlighted by Fig. \ref{fig:cdte}. 

To the authors' knowledge, multi-agent RL has been explored by two previous investigations. The first explored planning and management problems involving multiple independent entities operating with deterministic retailer demand and lead-times \citep{FujiMARLBDG}. The second work focused on the examination of MARL performance in serial supply chains, with backlog, dual-sourcing, and lost sales under stochastic demand with benchmark to commonly used heuristic re-order policies \citep{liu2022multi}. While MARL is powerful, the associated algorithms are characterized by many hyperparameters and different amounts of information sharing in policy identification. Further, MARL does not generally have the stability, safety and feasibility guarantees observed in traditional optimal control methods such as model predictive control (MPC) \citep{GORGES_MPC_RL}. This has slowed the adoption of MARL methods to control large real-world systems. 

\subsection{Motivation}

In this work, we provide a thorough interrogation of MARL methods and their performance in single-product, multi-echelon inventory control problems characterized by different production network structures, sizes, and sources of uncertainty (i.e. both demand and lead time uncertainty). We explore the effects of varying degrees of information centralization in policy identification and benchmark the resultant joint policy performance to a distributed optimization approach utilizing nominal descriptions of uncertain variables. In doing so, we develop on previous works that have provided preliminary but limited interrogation of these algorithms. We provide comments on the potential application of MARL methods in practice and scope for future work.

\section{Problem Description}
\label{sec: Problem Description}
\subsection{Mathematical Formulation}
\label{sec: mathematical formulation}
The IM problem can be formulated as a constrained optimization problem. The discrete-time formulation is considered here, where the aim of the optimization is to find the optimal re-order quantity at each node $m$ and each time period $t$ for all $M$ nodes over a fixed horizon of $T$ time periods. This can be written as:

\footnotesize
\begin{subequations}
\begin{align}
\begin{split}\label{eqn: optimisation}\scriptscriptstyle
&\max \sum_{m=1}^M \sum_{t=1}^T P^m s_r^m[t] - C^m o_r^m[t] - I^m i^m[t] - B^m b^m[t] \,, 
\end{split}\\
\begin{split}
&\text{subject to} \notag 
\end{split}\\
\begin{split}
\label{eqn: inventory}
&i^m[t] = i_0^m[t] - s_r^m[t] + a_r^m[t]\,, \quad \forall m, \forall t, 
\end{split}\\
\begin{split}
\label{eqn: backlog}
&b^{m_d}[t] = b_0^{m_d}[t] - s_r^{m_d}[t] +  d_r^{m_d}[t], \quad \forall m, \forall d \in \mathcal{D}_m\,,
\end{split}\\
\begin{split}
\label{eqn: sales constraint 2}
&s_r^{m_d}[t] \leq b_0^{m_d}[t] + d_r^{m_d}[t]\,, \quad \forall m, \forall t, \forall d \in \mathcal{D}_m 
\end{split}\\
\begin{split}
\label{eqn: sales constraint 1}
&s_r^m[t] \leq i_0^m[t] + a_r^m[t],  \quad \forall m, \forall t,    
\end{split}\\
\begin{split}
\label{eqn: acquisition}
&a_r^m[t] = s_r^{m_u}[t-\tau_r^m], \quad \forall m \neq 1\,, t\geq \tau_r^m  
\end{split}\\
\begin{split}
\label{eqn: factory}
&a_r^1[t] = o_r^1[t-\tau_r^1], \quad t\geq \tau_r^1
\end{split}\\
\begin{split}
\label{eqn: order demand equality}
&d_r^{m_d} = o_r^d,  \quad \forall m, \forall d \in \mathcal{D}_m\,,  
\end{split}\\
\begin{split}
&\text{with} \notag 
\end{split}\\
\begin{split}
\label{eqn: customer demand}
& d_r^{m}[t] = c^{m}[t], \quad \forall m \in \mathcal{C}, \forall t   
\end{split}\\
\begin{split}
\label{eqn: limits}
&o_r^m[t] \leq O_{r_{\max}}^m\,, i^m[t] \leq I_{\max}^m\,, \quad \forall m, \forall t,  
\end{split}
\end{align}
\end{subequations}
\normalsize

where $s_r$ is the sales/amount of goods shipped to a downstream node (or customers), $o_r$ is the replenishment order, $d_r$ is demand from downstream nodes, $i$ is the on-hand inventory at the end of a time period, while $b$ is the backlog at the end of a time period and $a_r$ is the acquisition at each node i.e. the goods received from an upstream node. $i_0$ and $b_0$ represent the on-hand inventory and backlog at the start of each period respectively. The parameters $P$ and $C$ are the price of the goods sold and the cost of replenishment orders respectively. $I$ and $B$ represent the storage and backlog costs respectively. $I_{\max}$ and $O_{r_{\max}}$ represent limits on node storage and replenishment order amount. The subscript $u$ refers to the upstream node such that $m_u$ is the upstream node of $m$, while the subscript $d$ refers to downstream nodes. Therefore the total backlog $b^m$ and shipment $s^m$ of a node $m$ to its downstream nodes is the summation of the backlog and shipment to each downstream node ($b^{m_d}$ and $s^{m_d}$) respectively. The set $\mathcal{D}_m$ is the set of direct downstream nodes of node $m$ while $\mathcal{C}$ is the set of nodes with customer demand where $c$ refers to customer demand.\par
Equation (\ref{eqn: optimisation}) represents the objective, which, in this case, is to maximize total profit across the entire SCN. The constraints (\ref{eqn: inventory}) and (\ref{eqn: backlog}) govern how the inventory and backlog update over time. Equations (\ref{eqn: sales constraint 2}) and (\ref{eqn: sales constraint 1}) are constraints on the amount of goods a node can ship downstream where it cannot exceed on-hand inventory or downstream demand and backlog. Equation (\ref{eqn: acquisition}) represents the lead time of a shipment where a good shipped to node $m$ will take $\tau_r^m$ periods to arrive at that downstream stage, while (\ref{eqn: factory}) refers to the acquisition of the root node that produces goods where the lead time expresses time to manufacture the goods. 

\subsection{Sources of uncertainty}
\label{sec: sources of uncertainty}
In this work, we consider exclusively exogenous forms of uncertainty on the customer demand at the retailer nodes, $d_r^{m}\,, \enspace \forall m \in \mathcal{C}$ , and on lead times of order delivery, $\tau_r^m \,, \enspace \forall m$.\par
We model the stochastic customer demand in two different ways. In general, we consider the demand uncertainty to be modeled via a stationary Poisson distribution with a constant rate parameter. However, customer demand is not always stable as large fluctuations can occur like a rush to buy products for example. We model these spikes in customer demand by modifying the stationary demand profile where the modification depends on two independent random binary variables with a Bernoulli distribution. At each time step the customer demand drawn from the Poisson distribution may be subject to a multiplier based on the outcome of the random variables $X_{t_0}$ and $X_{t_2}$. The outcome of $X_{t_0}$ represents whether the multiplier is $0$ while the outcome of $X_{t_2}$ represents if the multiplier is $2$ with a probability $p_{\text{d}}$.

Meanwhile, the lead time of products is approximated both as deterministic and stochastic depending on the computational experiment investigated. When described as stochastic, the arrival of delivery is drawn from a finite Bernoulli process, $X_{\tau_r^m} \ldots, X_{\tau_r^m+n}$, of binary random variables each described by a Bernoulli distribution. The outcomes of a constituent binary random variable, $X_t$, in the process represent whether the delivery has been made or not at a number of discrete time indices, $\textit{t}$, from the discrete-time index of order placement with a probability $p_{\tau}$. This also implies that the length of the Bernoulli process is built recursively until one of the binary random variables assigns delivery, at which point the Bernoulli process is fully constructed by $n$ binary random variables with $X_{\tau_r^m + n} = 1$ and $X_t=0$ where $\tau_r^m\leq t<\tau_r^m + n$. Intuitively, this scheme can be thought of as flipping a fair coin until heads is observed, at which point no further trials are conducted (i.e. the product has been delivered). This is a description similar to that detailed in \cite{gurnani1996optimal}.

With the two different methods of modeling customer demand and delivery lead times described above, we investigate different combinations of them in \S\ref{sec:experiments}. The different uncertainty setting combinations are summarised in Table \ref{tab: uncertainty conditions} below. Unless otherwise stated, the setting with a stationary Poisson distribution for customer demand and deterministic delivery lead times \textbf{S1} is used. The robustness of the methods proposed to non-stationary demand and stochastic lead-times, \textbf{S2} and \textbf{S3} respectively is investigated in \S\ref{sec: uncertainty}.

\begin{table}[H]
  \centering
  \caption{Customer demand and delivery lead time uncertainty settings.}
  \resizebox{0.95\textwidth}{!}{
    \begin{tabular}{ccc}
    \hline
    \textbf{Experimental Condition} & \textbf{Demand Description} & \textbf{Lead-time Description} \\
    \hline
    \textbf{S1} & Stationary Poission & Deterministic \\
    \textbf{S2} & Non-stationary modified Poisson & Deterministic \\
    \textbf{S3} & Stationary Poission & Uncertain Bernoulli Process \\
    \end{tabular}
    }%
  \label{tab: uncertainty conditions}%
\end{table}%


\subsection{Decentralized Partially Observable Markov Decision Process Formulation}
\label{sec: environment}
\begin{figure}
    \centering
    \includegraphics[width=0.63\linewidth]{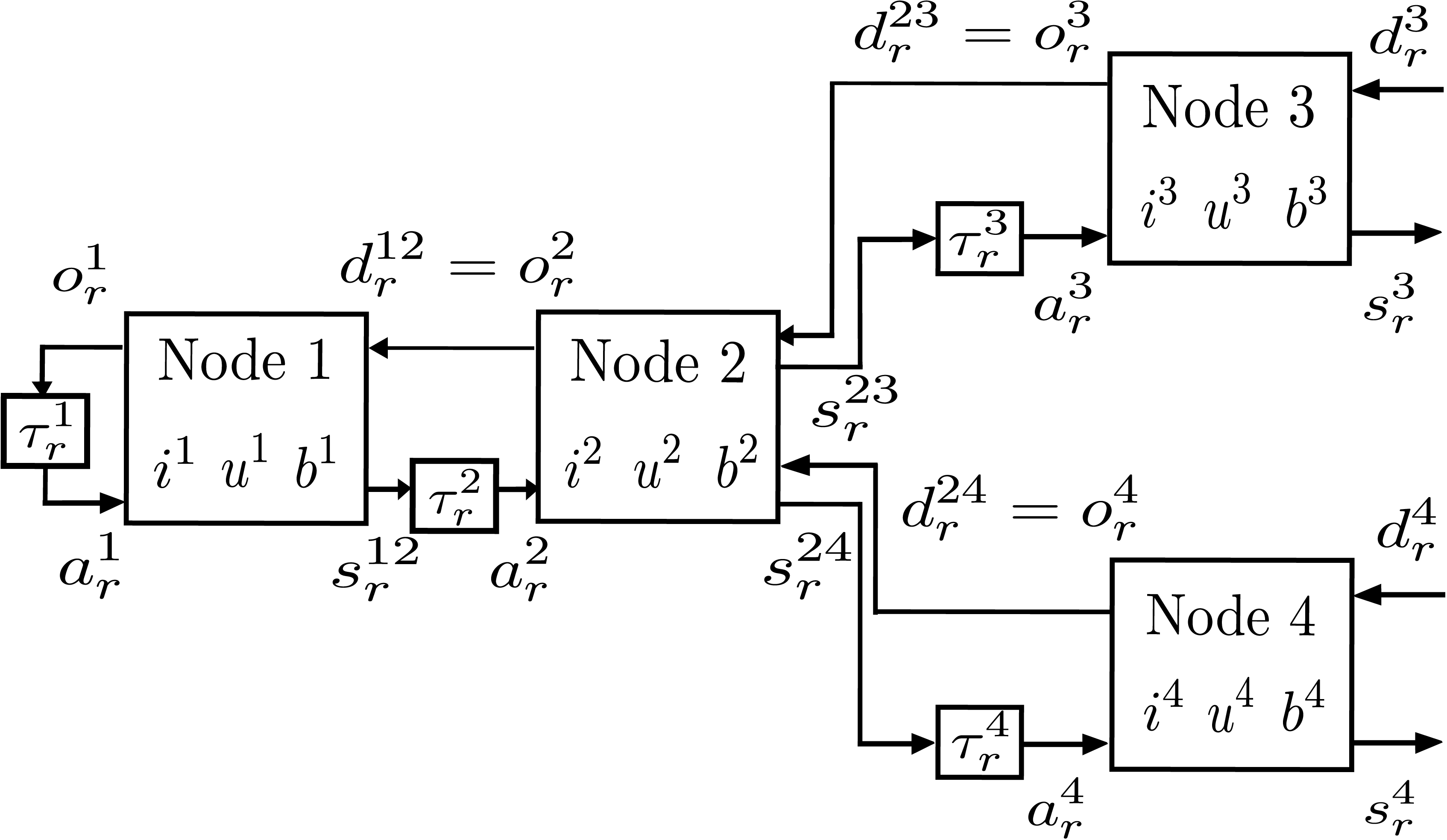}
    \caption{An idealized divergent multi-echelon supply chain network model with four nodes.}
    \label{fig:multi-echelon}
\end{figure}

In order to use reinforcement learning, the IM problem has to be formulated as a decentralized partially observable Markov decision process (Dec-POMDP) \citep{decPOMDP} which is a generalization of the Markov decision process (MDP) that considers decentralized agents, each with only partial observability of the full system state. It is formally defined as an 8-tuple $<\mathcal{N}, \mathcal{S}, \mathcal{A}, \mathcal{R},  \mathcal{P}, \Omega, \mathcal{O}, \gamma>$ where
\begin{itemize}
    \item $\mathcal{N}$ is the number of agents.
    \item $\mathcal{S}$ is the set of all valid states.
    \item $\mathcal{A}^j$ is the set of actions of agent \textit{j}. $\mathcal{A}$ denotes the set of all actions $\mathcal{A}:=\mathcal{A}^1 \times \dots \times \mathcal{A}^\mathcal{N}$.
    \item $\mathcal{R}^j: \mathcal{S} \times \mathcal{A} \times \mathcal{S} \to \mathbb{R}$ is the reward for agent $j$ for a transition from ($\textbf{s}$,$\mathbf{a}$) to $\textbf{s}'$.
    \item $\mathcal{P}: \mathcal{S} \times \mathcal{A} \to \mathbb{P}(\textbf{s})$ is the state transition function.
    \item $\Omega_j$ is the set of observations for agent $j$, with $\Omega:=\Omega^1 \times \dots \times \Omega^\mathcal{N}$.
    \item $\mathcal{O}$: $\mathcal{S} \times \mathcal{A} \rightarrow \mathbb{P}(\textbf{o})$ is a set of conditional observation probabilities $\mathbb{P}(\textbf{o}|\textbf{s},\textbf{a})$
    \item $\gamma$ is a discount factor for future rewards.
\end{itemize}
An environment representing the IM problem considered was built to train RL agents to find the optimal re-ordering policy. A visualization of the environment for a divergent SCN is shown in Figure \ref{fig:multi-echelon} for example.

Since the multi-period variant of the IM problem was considered, each discrete time period is represented by a time-step in the environment where each episode has a fixed length of $T$ time-steps. At each step, when the agents take their actions, the transition dynamics of the system can be described by the following sequence.
\begin{enumerate}
    \item All nodes place their replenishment order, $o_r^m$, to their upstream node except the root node where the replenishment order triggers the production of the good.
    
    \item Nodes receive the goods shipped by their upstream node after a given lead time $\tau_r^m$. The lead time can be deterministic or stochastic based on the experimental condition used, as described in \S\ref{sec: sources of uncertainty}. The acquisition is zero for each node $m$ when $t < \tau_r^m$.
    
    \item All nodes not in $\mathcal{C}$ receive a demand $d^{m_d}$ from each of their downstream nodes, (equal to their replenishment order), while nodes in $\mathcal{C}$ receive customer demand.
    
    \item This demand is fulfilled along with any backlog $b^{m_d}$ from the available inventory at each node, $i^m$ as well as any goods received from an upstream node, $a^m_r$. Fulfillment priority is given to the existing backlog.
    
    \item Any demand from a downstream node that is unfulfilled is added to the node's existing backlog. All backlog at the end of the time period incurs a cost. The backlog update is given by
    \begin{equation}
        b^m[t+1] = b^m[t] + \sum_{d \in \mathcal{D}_m} d^{m_d}[t] - \sum_{d \in \mathcal{D}_m} s_r^{m_d}[t].
    \end{equation}
    
    \item The inventory at the end of the time period at each node is also updated after sales and acquisition with the following update rule
    \begin{equation}
        i^m[t+1] = i^m[t] + a_r^m[t] - \sum_{d \in \mathcal{D}_m} s_r^{m_d}[t]
    \end{equation}
    This is surplus inventory that is held at a cost.
    
    \item Finally, the pipeline inventory/unfulfilled orders $u^m$ at the end of the time period is updated using
    \begin{equation}
    u^m[t+1] = u^m[t] + o_r^m[t] - a_r^m[t]
    \end{equation}
    
\end{enumerate}

It is worth noting that the environment was built using the OpenAI Gym framework \citep{openai_Gym}.

\section{Multi-Agent Reinforcement Learning}
\label{sec: RL Agents}
Reinforcement learning has been shown to learn re-ordering policies for the multi-echelon IM problem that are competitive with methods used in industry \cite{hubbs2020orgym, ORgymextension}. However, learning the optimal policy becomes more difficult for a centralized agent as the problem space becomes larger for larger SCNs. More importantly, SCNs are made up of separate entities with limited information sharing that would make any centralized solution infeasible. Since the multi-echelon IM problem can be naturally decomposed into a multi-agent system where each node is controlled by an agent, MARL can help overcome these constraints. \par

MARL algorithms can however have varying levels of decentralization. In \cite{dynamics_of_coop_MARL}, the authors classify MARL methods as being either \emph{joint action learners} (JALs) or \emph{independent learners} (ILs). JALs can observe the actions of all other agents while ILs can observe only their actions. The limited observability of ILs makes the environment non-stationary from the agent's perspective, which may lead to sub-optimal policies. While JALs do not have that limitation, they are not practical in most real-world applications due to the agents being physically unable to access the actions and observations of other agents in the environment. However, with access to a centralized simulation environment, agents could be allowed to view the actions and observations of other agents during training only and yet execute actions in a decentralized manner allowing the agents to be deployed separately as with the \emph{centralized training decentralized execution} (CTDE) framework detailed by \cite{CTDE}. Both independent and joint action learners are investigated by leveraging different MARL algorithms with varying levels of information sharing.\par

\subsection{Algorithms}
In order to find the distributed optimal re-order policy, multi-agent variations of Proximal Policy Optimization \citep{PPO} were investigated. PPO has been successfully applied in many control problems in recent years and is simple to tune relative to other RL algorithms. The particular PPO algorithm used in this work utilizes an actor-critic approach and is discussed further in \ref{sec: Algorithm}. Three distinct multi-agent implementations of PPO were investigated in this work, each with varying levels of decentralization. \par

\begin{figure}
\centering
\begin{subfigure}{.48\textwidth}
\centering
\includegraphics[width=0.8\linewidth]{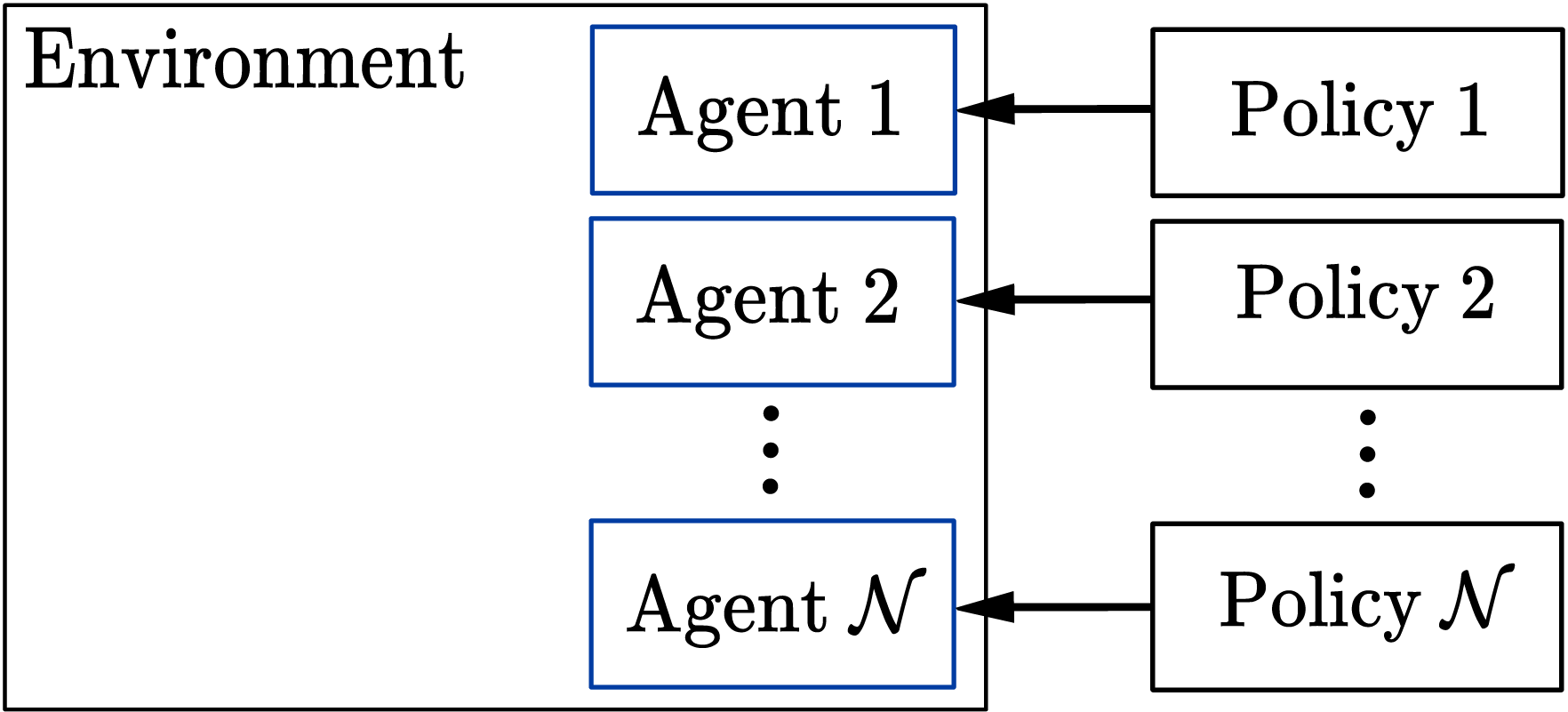}
\caption{Independent policies.}
\label{fig: independent policies}
\end{subfigure}%
\begin{subfigure}{.48\textwidth}
\centering
\includegraphics[width=0.8\linewidth]{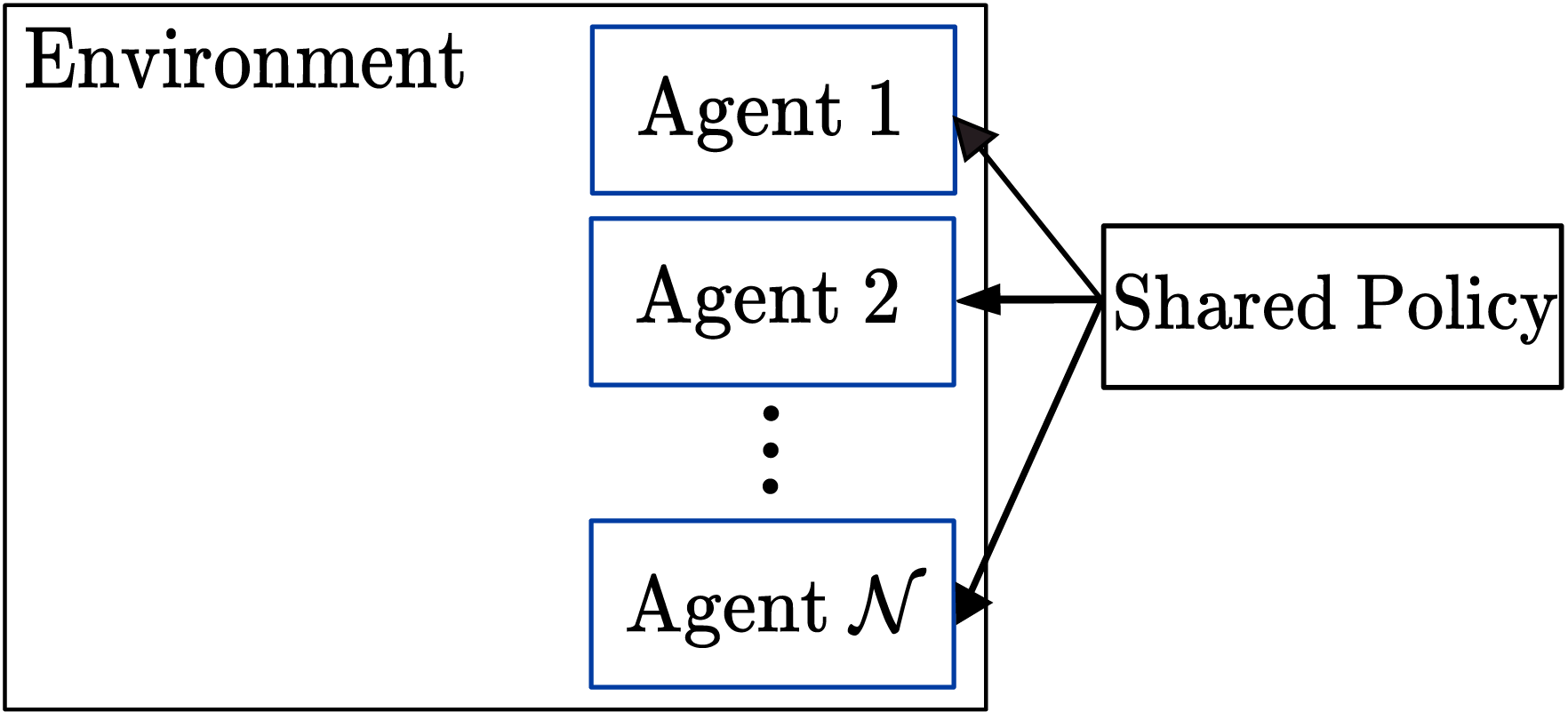}
\caption{Shared Policy.}
\label{fig: shared policies}
\end{subfigure}
\caption{The difference between agents with independent policies and parameter sharing.}
\label{fig: independent and shared policies}
\end{figure}

The first two are variations of \emph{Independent Proximal Policy Optimization} (IPPO) \citep{IPPO}. IPPO simply involves independent agents, each utilizing PPO in order to maximize a shared team reward. Given the use of a shared reward it is important to highlight the difference between the single agent implementation is that each agent acts according to observation of its own local state information and follows its own unique policy gradient in parameter updates. Previous results report that this form of IL matches or outperforms JALs such as QMIX \citep{QMIX} on benchmark maps of the Starcraft multi-agent challenge \citep{smac}. Furthermore, the authors argue that ILs such as IPPO scale better than JALs with the number of agents as they do not face the issue of a growing observation space. The two implementations of IPPO we investigate vary in the degree of information sharing for the formation of each agent's policy gradient and therefore in the degree of decentralization. In the first implementation, each agent, controlling node $m$, uses and updates its own independent policy $\pi_{\theta_{m}}$ and will be referred to as \emph{IPPO} in this paper. The second, which will be referred to as \emph{IPPO with shared network}, is similar to the original implementation where agents share the same network parameters and thus update a single policy $\pi_{\theta}$ and sample actions conditioned on their own observation. The observation is augmented to include a unique identifier for each agent as an additional input. The difference between the two implementations is highlighted in Figure \ref{fig: independent and shared policies}.\par

The third method investigated is a JAL implementation of PPO called \emph{Multi-agent Proximal Policy Optimization} (MAPPO) first proposed in \cite{MAPPO} and was shown to achieve state-of-the-art performance. In this method, each agent has an independent actor-network that takes as input the agent's observation only, while utilising a central critic network. The central critic network can be shared amongst all agents, taking as input all agent observations and actions. Another implementation of the centralized critic involves each agent having their own critic network that takes as input the agent's own observation as well as the observations and actions of all other agents. The latter implementation was used in this work. While centralized training is required to provide the critic with the actions of other agents, each agent can still execute actions in a decentralized manner as the actor network takes only the agent's own local observations as input. The use of a centralized critic during training makes the environment more stationary from the agent's perspective as it can observe all other agents' actions. Training is improved as each agent has access to better estimates of the value function $V(\textbf{s})$. This ultimately provides more stable learning dynamics in the identification of the optimal joint policy.\par

\subsection{Neural Network Architecture}
\begin{figure}
\centering
\begin{subfigure}{.35\textwidth}
\centering
\includegraphics[width=0.85\linewidth]{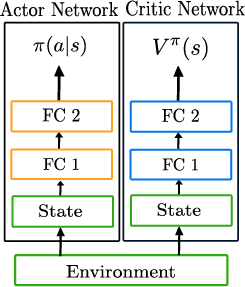}
\caption{Network without RNN.}
\label{fig: network no rnn}
\end{subfigure}%
\begin{subfigure}{.35\textwidth}
\centering
\includegraphics[width=0.85\linewidth]{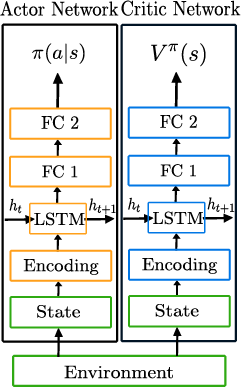}
\caption{Network with RNN.}
\label{fig: network rnn}
\end{subfigure}
\caption{The neural network architecture, composed of fully connected (FC) hidden layers, used by the RL agents.}
\label{fig: Single agent Network architecture}
\end{figure}

An actor-critic architecture was utilized where each agent (or set of agents in the case of shared policies) use two neural networks. The critic network is used to estimate the state-action value function $Q(s, a)$, parameterized by parameters $\phi$, while the policy $\pi$ is parameterized by an actor network with parameters $\theta$.
The general structure of the networks used in this work is shown in Figure \ref{fig: Single agent Network architecture}.\par

 Two different neural network architectures were investigated to assess whether the use of \emph{recurrent neural network} (RNN) layers would lead to better re-order policies. This was motivated by time-dependencies in the inventory management problem and partial observability of individual node agents which result in an environment with non-Markovian characteristics. In \cite{reccurrentpolicygradient} the authors propose the \emph{recurrent policy gradient}, which is a policy gradient method that utilizes RNNs, particularly Long Short-term Memory layers \citep{LSTM}. It was argued that by combining policy gradient methods with \emph{back-propagation through time} \citep{BPTT}, one can improve the performance of RL agents in \emph{partially observable MDPs} (POMDPs), where the agent does not have full state observability. In cases where the current state does not capture the history of the system, RNNs would alleviate this issue through the use of their hidden states $h_t$, which would capture the history of the system. For example, the presence of lead times $\tau$, means the true rewards associated with a particular action are not immediate, therefore contextualizing on history may help remedy the breakdown in the Markovian problem structure. \par

 The first neural network architecture considered consists of a multi-layer perceptron architecture with two fully connected hidden layers as shown in Figure \ref{fig: network no rnn} (denoted as FC1 and FC2 in the figure). The second architecture resembles the first with an additional RNN layer as shown in Figure \ref{fig: network rnn}. In the case of MAPPO, the critic network was modified to take the agent's own local observation as well as the observations and actions of all other agents that occurred in the same time step. While having a different architecture for the centralized critic and actor networks may be beneficial, this was not explored here.\par

Agents were trained using both neural network architectures for all the MARL algorithms we investigated in a four-stage serial supply chain to allow for comparison between them. While it was hypothesized that the use of an RNN-based network would significantly improve the performance of the multi-agent system due to each agent's limited observability, they were found to provide little to no performance improvement for all the MARL algorithms. Furthermore, they resulted in significantly longer training time and a higher deviation of rewards. Therefore the neural network architecture shown in Figure \ref{fig: network no rnn} was used for the RL algorithms in all the experiments in \S\ref{sec:experiments}. This analysis is described in further detail in \ref{sec: Agent Training}.

\subsection{Formulation}
We describe the state and action spaces of the MARL agents for the dec-POMDP described in \S\ref{sec: environment}. In the multi-agent version of the IM problem considered in this work, each node, $m$, in the SCN is controlled by a separate agent, $j$. Each agent gets an observation, which extends to variables associated with the node it controls only and decides the optimal action, re-order amount, for the node. 

\textbf{Action Space} The action each agent takes is the replenishment order at the node controlled by the agent. While the goods are traded in integer quantities as described in \S\ref{sec: environment}, a continuous action space was used. This was deemed a reasonable assumption given the requirement for the quantities to be integer. However, if the potential space of reorders was discretized more coarsely this assumption may demand revision. The action space of the policy was set to lie in the interval $[-1, \hspace{1mm} 1]$ as recommended by \cite{stable-baselines3}. The output actions of the agent were then re-scaled to the physical replenishment order range of $[0, \hspace{1mm} O^m_{r_{\text{max}}}]$ and rounded to the nearest integer.\par

\textbf{Rewards} The main objective of the IM problem in this work is to maximize the total profit in the whole SCN which would require agents to collaborate in order to achieve a common goal. Therefore the agents share the overall total reward such that the reward received by each agent at every time-step from the environment is
\begin{equation}
    \frac{1}{M}\sum_{m=1}^M P^m s_r^m[t] - C^m o_r^m[t] - I^m i^m[t+1] - B^m b^m[t+1],
    \label{eqn: multi agent reward}
\end{equation}
which is simply the mean of the profit earned by each node and therefore agent. Therefore actions taken that hurt overall profit will result in a lower reward for the agent, even if it increases the agent's node's profit.\par

\textbf{State Space} RL algorithms rely on the Markov assumption, which means the information contained within the state is sufficient to decide the action to take. This would lead to the conclusion that the more information captured by the state in each time step, the more this Markovian assumption is satisfied, leading to successful agent learning. However, unnecessary information may make it harder to learn the underlying system dynamics and therefore a good policy. It was therefore investigated which environment variables described in \S\ref{sec: environment} should be added to the state vector.\par

\begin{figure}[H]
    \centering
    \includegraphics[width=0.75\linewidth]{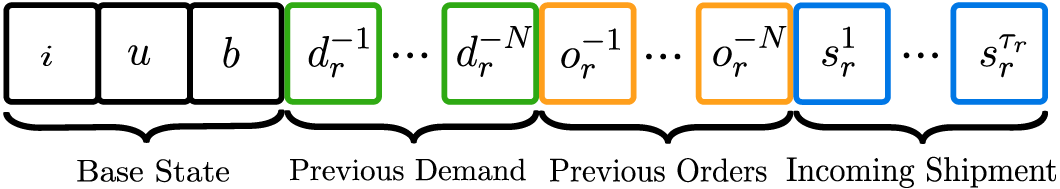}
    \caption{Node state representation configuration used by the RL agent.}
    \label{fig: state}
\end{figure}

The amount of on-hand inventory $i$, off-hand or pipeline inventory $u$ and backlog $b$ at each stage at the start of the time period were identified as crucial variables (the \emph{base state}) since they describe the order and material flow throughout the SCN. Other variables identified as potentially important are the demand and actions in the  $N$ previous time steps and the incoming shipment from an upstream stage. The incoming shipment variable is essentially a vector of size $\tau_m$ where each element represents the amount of goods being shipped and the index represents how many time-steps away it is from arriving at the node. It should be highlighted that if lead time is to be described as a random variable, then in general this part of the state is partially observed and can only be estimated via a forecast. This state vector is illustrated in Figure \ref{fig: state} below where the superscript denotes the time-step of each variable. \par

For all the MARL algorithms considered, agents were trained with different combinations of the variable groups in Figure \ref{fig: state} in a four-stage serial supply chain. The mean episode rewards of the trained agents were compared to assess which combination of variables in the agents' observations lead to the best performance for each algorithm. This analysis was done for each algorithm given the varying levels of information sharing between them which may affect what information is useful to include in the agents' observations. The hyperparameters of the algorithms were tuned for each combination of state variables to ensure a fair comparison between them. This analysis is described in further detail in \ref{sec: Agent Training}.\par

Following this analysis, it was found that the best performance for both IPPO algorithms was obtained with a state vector containing previous demand and incoming shipments in addition to the base state. While for MAPPO the best performance was achieved with all variable groups. In both cases the optimal of previous time-steps $N$ was found to be 1. These state vectors are used for numerical simulations in \S\ref{sec:experiments}.

\section{Experiments}
\label{sec:experiments}
The performance of the different MARL agents is investigated in different configurations of the IM problem with the aim of assessing which MARL method performs best. We also investigate how our MARL implementations compare against the distributed shrinking horizon LP-based method (DSHLP) described in \S\ref{sec: OR methods} as well as a centralized RL agent. \par

The centralized single agent has full observability, where the agent's state consists of the observations of all nodes. The agent's action is the re-order amount for each of the nodes in the SCN. The agent was trained using the same way as the MARL agents as described in \ref{sec: Agent Training}. The hyperparameters of the centralized RL agent as well as the MARL agents used to obtain the results below, are detailed in \ref{sec: Hyperparameter Values}.\par

All algorithms are compared against the performance of an \emph{Oracle}, which solves the IM problem as a deterministic problem with all customer demand known a priori using centralized LP.\par

\subsection{Base Case: Four-stage Chain}
\label{sec: base case}
This was the base case considered during the implementation of the RL methods due to its resemblance to the original beer distribution game \citep{BeerGame} and for which the hyperparameter values of the algorithms used were tuned. The particular configuration used is presented in Table \ref{tab:four stage config}. In order to assess the algorithms, we observe the performance of the trained agents on 200 simulated test episodes with 30 time-steps each and customer demand following a Poisson distribution with $\lambda_{\text{Poisson}} = 5$. The same test settings were used throughout all simulations described in this section.\par
The metrics used for comparison were the mean episode reward, inventory and backlog. Inventory and backlog refer to the mean inventory and backlog at the end of each time step in the entire SCN over an entire episode. The main metric of interest is the mean episode reward, however, the other metrics help provide intuition behind the performance achieved by each of the policies and how close their performance is to the optimal performance i.e. the Oracle. The results are presented in Table \ref{tab:four-stage results}.\par
%
\begin{table}[H]
  \centering
  \caption{Average performance metrics of the RL and LP algorithms for test episodes in a four-stage serial supply chain network.}
  \resizebox{0.55\textwidth}{!}{
    \begin{tabular}{lcccc}
    \hline
    \textbf{Method} & \textbf{Reward} & \textbf{Oracle} & \textbf{Inventory} & \textbf{Backlog} \\
    \hline
    \textbf{Oracle} & 619.4 & 1.00 & 49.3 & 2.7 \\
    \textbf{DSHLP} & 416.8 & 0.67  & 75.8  & 160.0  \\
    \textbf{Single Agent} & 477.1 & 0.77  & 183.4 & 84.0  \\
    \textbf{IPPO} & 443.8 & 0.71  & 317.9 & 64.8 \\
    \textbf{IPPO shared} & 439.4 & 0.71 & 263.2 & 100.1 \\
    \textbf{MAPPO} & 463.0 & 0.75 & 222.7 & 81.0 \\
    \end{tabular}
    }%
  \label{tab:four-stage results}%
\end{table}%

\begin{figure}[H]
    \centering
    \includegraphics[width=0.75\linewidth]{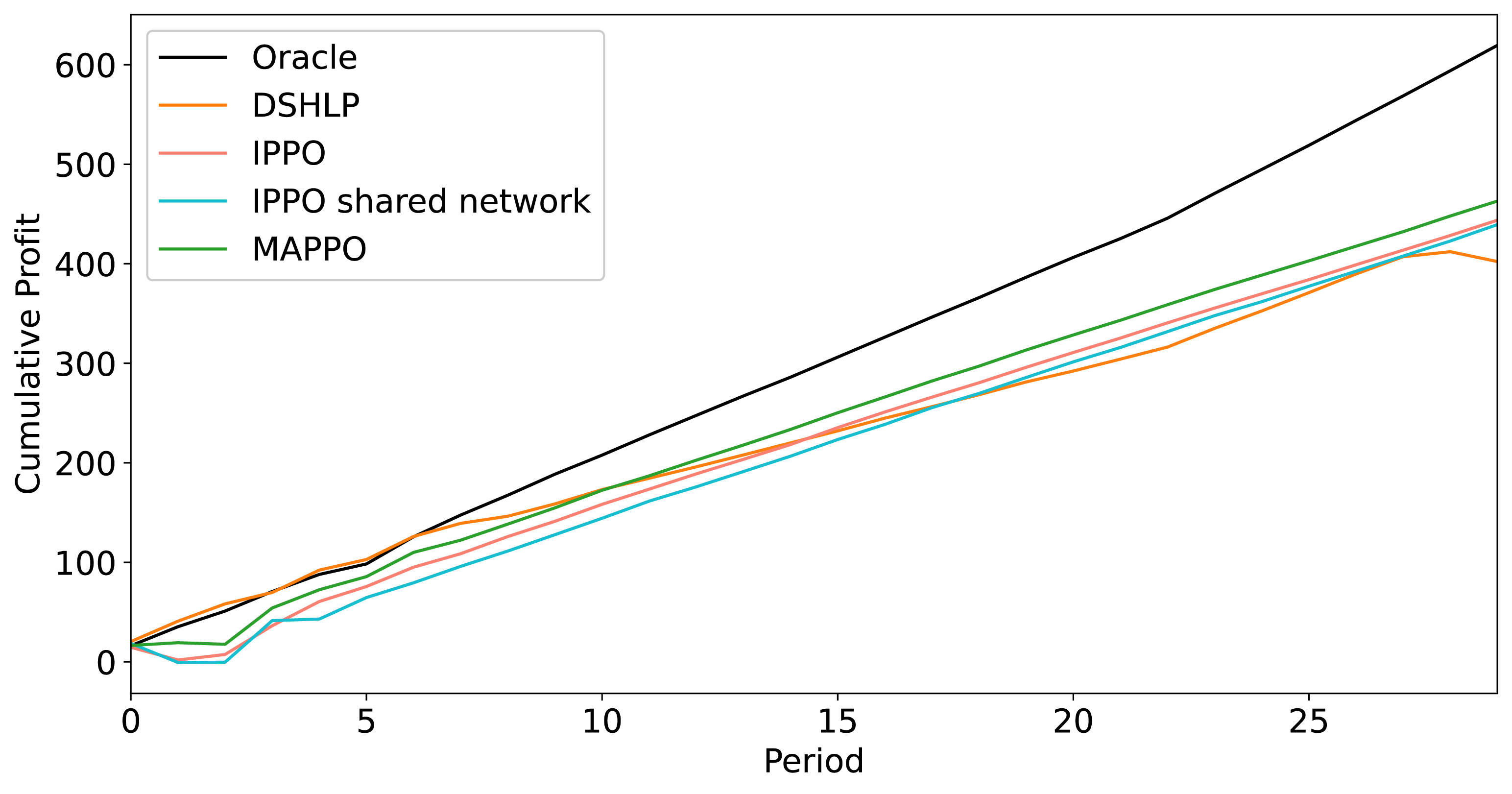}
    \caption{The average cumulative profit achieved by the multi-agent RL algorithms in a four-stage supply chain network compared with LP implementations.}
    \label{fig: 4 stage cumulative profit}
\end{figure}

It can be seen that all MARL methods outperform DSHLP in terms of the mean reward achieved. While the MARL algorithms seem to have a larger level of inventory on average than DSHLP, they had a much lower level of backlog on average. Of the MARL algorithms considered, the JAL MAPPO algorithm has achieved the best performance. The JAL MAPPO outperforms both ILs, which have a similar performance. This indicates there is an advantage to ensuring observability during state-value function learning - allowing for better-performing agents that are better able to coordinate. While the centralized RL agent performs better than all the MARL algorithms, its mean reward is marginally better than that achieved by the MAPPO agent, where MAPPO achieves only two percentage points less than the centralized agent. We can observe the average performance of each distributed algorithm across an episode in Figure \ref{fig: 4 stage cumulative profit} where we show the average cumulative profit across time steps.
One can see that while DSHLP initially performs better, all MARL algorithms eventually generate more profit with MAPPO consistently outperforming the other two MARL methods. While they generate less absolute profits than the Oracle, as is expected, the profit generated by all MARL agents across episodes follows the same trend as the Oracle.

\subsection{Divergent Supply Chain Network}
\label{sec: divergent chain}
The performance of the MARL methods is investigated for a divergent SCN where the network shown in Figure \ref{fig:multi-echelon} is considered with the configuration in Table \ref{tab:div config}. Divergent SCNs require more coordination as there is more than one source of external uncertainty due to the presence of multiple nodes with customer demand and divergent nodes are required to fulfil the demand of multiple nodes simultaneously. The dynamics of the system are also slightly different since nodes have to fulfill backlogs before current demand, which could make lead times more difficult to learn for nodes that share an upstream node. The performance of the different algorithms for the SCN is shown in Table \ref{tab:div results}.
\begin{table}[H]
  \centering
  \caption{Average performance metrics of the RL and LP algorithms for test episodes in a divergent supply chain network.}
  \resizebox{0.55\textwidth}{!}{
    \begin{tabular}{lcccc}
    \hline
    \textbf{Method} & \textbf{Reward} & \textbf{Oracle} & \textbf{Inventory} & \textbf{Backlog} \\
    \hline
    \textbf{Oracle} & 926.3 & 1.00 & 26.8 & 4.0 \\
    \textbf{DSHLP} & 754.3 & 0.81  & 160.8  & 159.1  \\
    \textbf{Single Agent} & 716.1 & 0.77  & 265.2 & 116.3  \\
    \textbf{IPPO} & 642.3 & 0.69  & 356.2 & 171.4 \\
    \textbf{IPPO shared} & 563.3 & 0.61 & 542.3 & 122.4 \\
    \textbf{MAPPO} & 651.2 & 0.70 & 443.5 & 96.8 \\
    \end{tabular}
    }%
  \label{tab:div results}%
\end{table}%

The DSHLP algorithm outperforms all MARL algorithms for this divergent SCN with a 10 percentage point difference (in terms of the optimal outcome). DSHLP also outperforms the centralized RL, which we include for reference, showing that data-driven methods might not handle divergent SCNs as well as serial SCNs and might not necessarily outperform traditional model-based optimization methods in all cases. That being said, using a higher network capacity and retuning algorithm hyperparameters (as discussed in Section \ref{sec: Algorithm}) for the divergent SCN might lead to better MARL performance due to the increased complexity of this type of chain. Furthermore, MAPPO and IPPO still learn a decent policy while having an execution speed that is orders higher than DSHLP.\par
As with the four-stage chain, MAPPO outperforms the other IPPO algorithms, however, its performance is close to IPPO. Unlike previously, using a shared network with IPPO leads to much worse performance. This could be due to the fact that different nodes experience slightly different dynamics in the divergent SCN as opposed to nodes in a serial SCN. For example, the policy of the divergent \emph{Node 2} with two downstream nodes, will be different from the other nodes, therefore using a shared network will lead to worse overall performance.  Using a shared network therefore might be limited to scenarios where all nodes in a SCN experience the same dynamics.

\subsection{Uncertainty}
\label{sec: uncertainty}
The performance of the different algorithms to disturbances in customer demand and lead time uncertainty were investigated to assess their robustness for the serial four-stage SCN described in \S\ref{sec: base case}. In the analysis below we show the performance of the different MARL algorithms re-trained at the different levels of uncertainty, as described in \S\ref{sec: sources of uncertainty}.

\begin{figure}
\centering
\begin{subfigure}{.49\textwidth}
\centering
\includegraphics[width=0.99\linewidth]{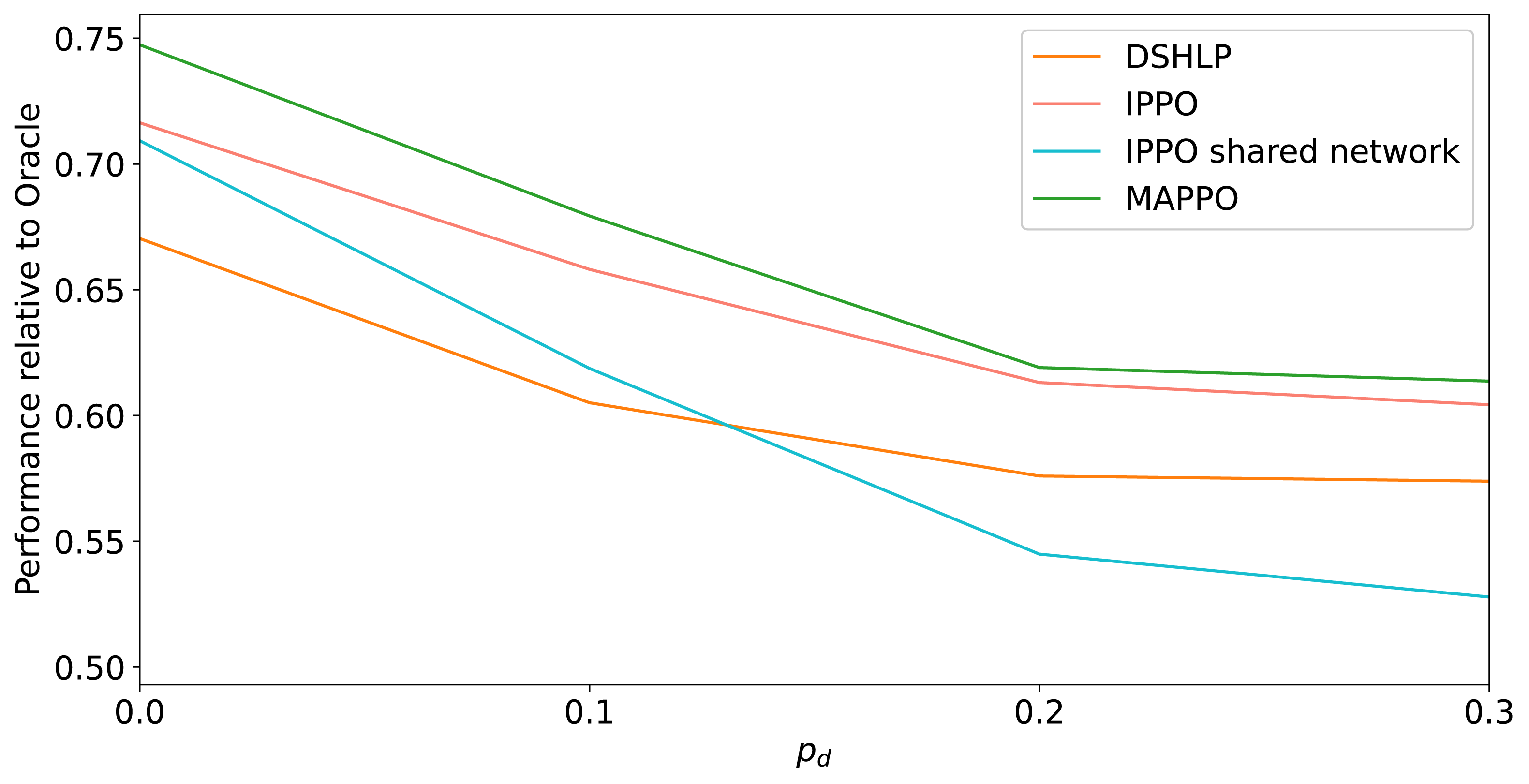}
\caption{Effect of uncertain customer demand.}
\label{fig:demand noise}
\end{subfigure}%
\begin{subfigure}{.49\textwidth}
\centering
\includegraphics[width=0.99\linewidth]{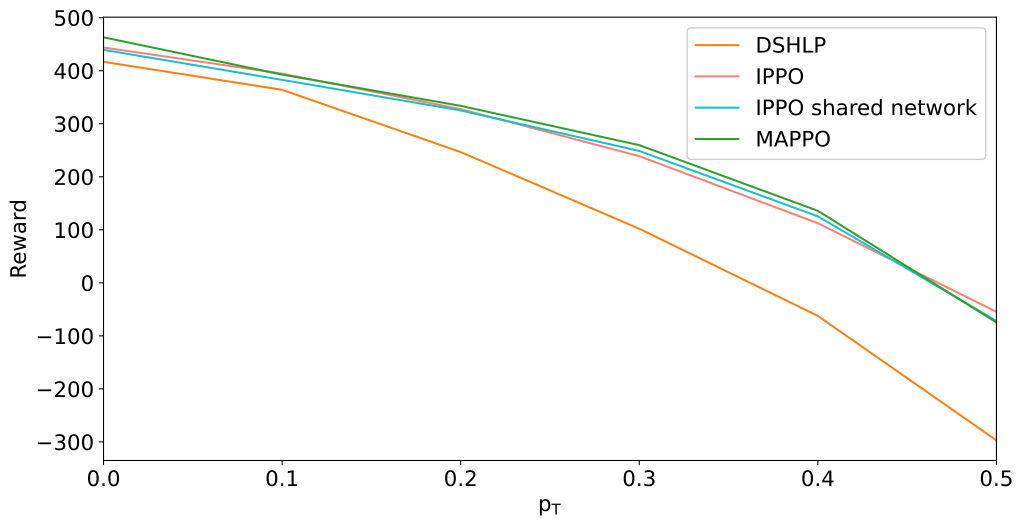}
\caption{Effect of uncertain lead times.}
\label{fig:delay noise}
\end{subfigure}
\caption{Effect of uncertainty in system parameters on the performance of the multi-agent RL algorithms.}
\label{fig: noise effect}
\end{figure}

\textbf{Customer Demand} While customer demand is uncertain, it still follows a known distribution. This is explicitly modeled in the model-based optimization method but learned by the RL methods from observations of different realizations of demand during training. In order to investigate the robustness of the different algorithms to non-stationary customer demand we use the \textbf{S2} setting for uncertainty as described in \S\ref{sec: sources of uncertainty}. The mean reward relative to the Oracle achieved by the different algorithms at 3 different levels of $p_{d}$ is shown in Figure \ref{fig:demand noise}. 
 MARL methods perform better in terms of absolute reward with MAPPO and IPPO both outperforming the distributed LP method. IPPO seems to be the most robust among the MARL methods experiencing the smallest drops in performance, however, MAPPO still outperforms IPPO in absolute terms across all levels of $p_{d}$ tested and experiences marginally larger drops in performance. Using a shared network with IPPO however, leads to a large drop in performance which shows that sharing experience across agents using the same network is less robust.\par
 While it was expected the model-free methods would be more robust to spikes in customer demand due to not explicitly assuming the distribution it is drawn from as well as being trained on this noisy demand, they seem less robust to that disturbance than the traditional distributed LP method. It can be seen that the distributed LP method experiences smaller drops in performance relative to the base case than the MARL methods, which might indicate that it is more robust to that type of uncertainty.

\textbf{Lead Times} Fixed lead times are explicitly modeled in the LP method as a constraint in the optimization problem of the system as in (\ref{eqn: acquisition}), but it is only learned from episodes of experience by the RL agents. We investigate the robustness of the different distributed algorithms to stochastic delivery lead times using the uncertainty setting \textbf{S3} as described in \S\ref{sec: sources of uncertainty}. The mean reward achieved by the different algorithms is shown in Figure \ref{fig:delay noise}.

It can be seen that the performance of the MARL is less affected by the stochastic delivery lead times relative to the distributed LP method. They outperform the distributed LP method significantly at levels of $p_{\tau}$ as the model-based method experiences a severe drop in performance. This large performance difference could be attributed to the fact that the model-based LP method explicitly models deterministic lead times and therefore experiences this large drop in performance. Another reason for MARL methods' outperformance could be their higher inventory levels as seen in Tables \ref{tab:four-stage results} and \ref{tab:div results} which makes them more robust to lead time uncertainty. This is in contrast to the stochastic customer demand where the distribution is taken into account, thus it is more robust to customer demand uncertainty. All MARL methods seem to have very similar performance all experiencing roughly the same drop in performance. MAPPO seems to experience the largest drop in performance while IPPO experiences to smallest, however, the difference is negligible.

\subsection{Supply Chain Network Size}
\label{sec: problem size}
The effect of the size of the SCN on the performance of the different RL algorithms presented is investigated. The performance of the algorithms was assessed in three serial SCNs with a different number of stages (two, four and eight with configurations for two and eight-stage SCNs in Tables \ref{tab:two stage config} and \ref{tab:eight stage config} respectively). The agents of each algorithm were trained in the new environments, however using the same hyperparameter values optimized for the four-stage SCN in \S\ref{sec: base case}. The mean and standard deviations of rewards achieved by each algorithm over 200 test episodes normalized by the Oracle's mean reward are shown below in Figure \ref{fig: stage comparison} for the different SCNs.\par

\begin{figure}[h]
    \centering
    \includegraphics[width=0.75\linewidth]{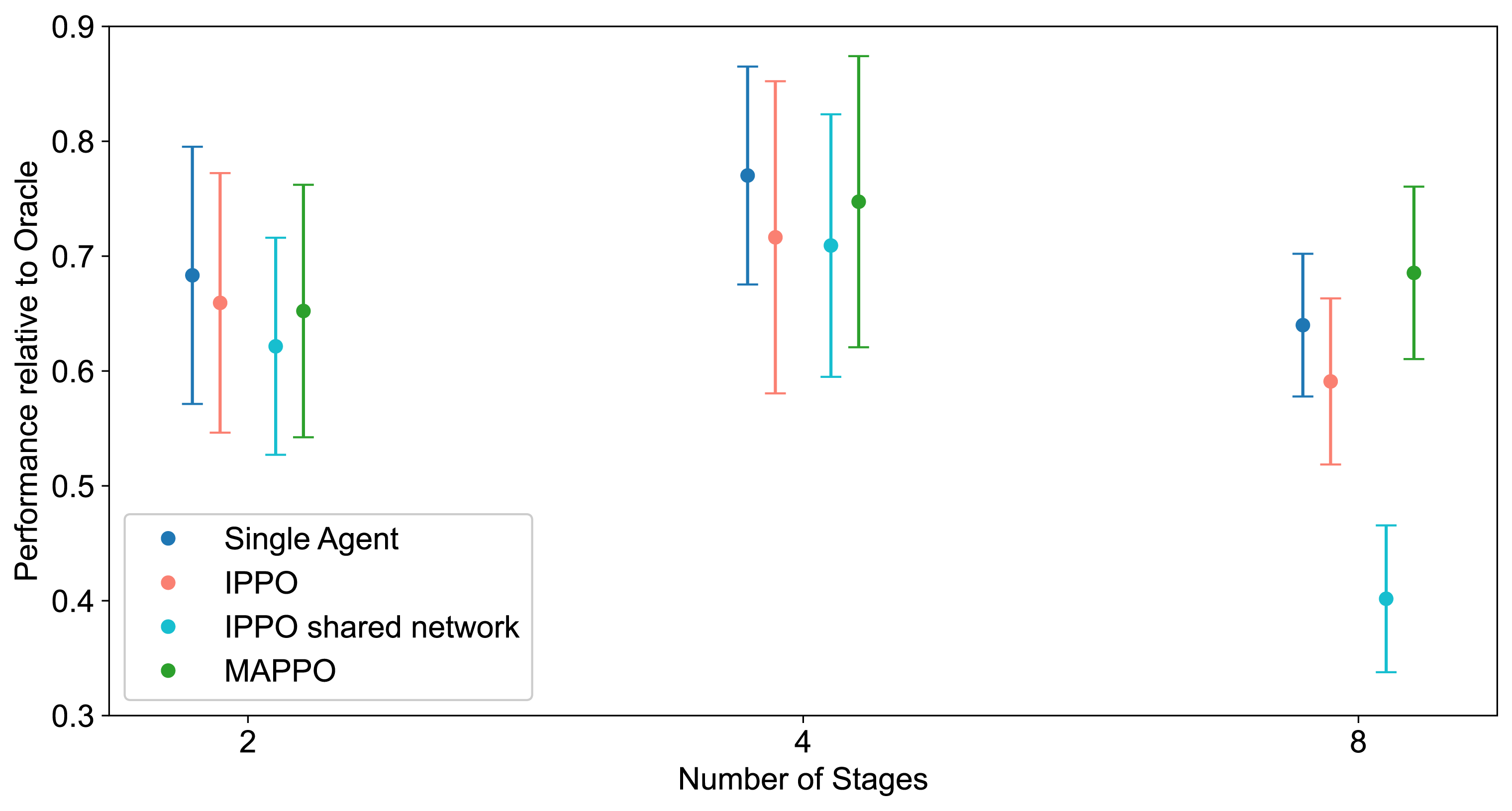}
    \caption{Effect of supply chain network size on the performance of the multi-agent RL algorithms, relative to a perfect information centralized LP implementation.}
    \label{fig: stage comparison}
\end{figure}

The MARL methods seem to be more robust to changes in the size of the problem. While the single agent outperforms the MARL methods in most cases when measuring absolute performance, its performance relative to the base four-stage case is worse than the MARL algorithms (with the exception of IPPO with a shared network). This can be seen by observing the change in performance from the base case. MAPPO even outperforms the single agent in the eight-stage SCN, while IPPO performs similarly to the single agent. A likely reason for this is that as the dimensionality of the SCN increases, the optimal policy function may no longer lie in the space of functions that can be expressed by the policy function structure of the single agent, suggesting a benefit to exploiting the problem structure through decentralization. The performance of IPPO with a shared network also suffers for the larger eight-stage SCN for possibly the same reason, as the shared network may not have sufficient capacity for providing a good policy for a larger number of agents. \par
It should be noted that if re-optimization of the single agent's hyperparameters had been considered the single agent policy may have not been subject to such a result with increasing supply chain size. The single agent is more affected by the problem's size than the MARL algorithms as both the state and action spaces of the problem grow, therefore may benefit from a different policy function approximation or different hyperparameter values. However, for agents in MARL algorithms, the size of the SCN does not affect a node's individual sub-problem therefore the action and state spaces stay constant. Because of this, MARL algorithms are more robust to changes in the problem's size, regardless of hyperparameter values.  This is a desirable attribute as the structure of SCNs can change over time and this result indicates a high-performance MARL algorithm can be transferred across different problems.

\subsection{Training without Noise}
\begin{figure}
\centering
\begin{subfigure}{.49\textwidth}
\centering
\includegraphics[width=0.99\linewidth]{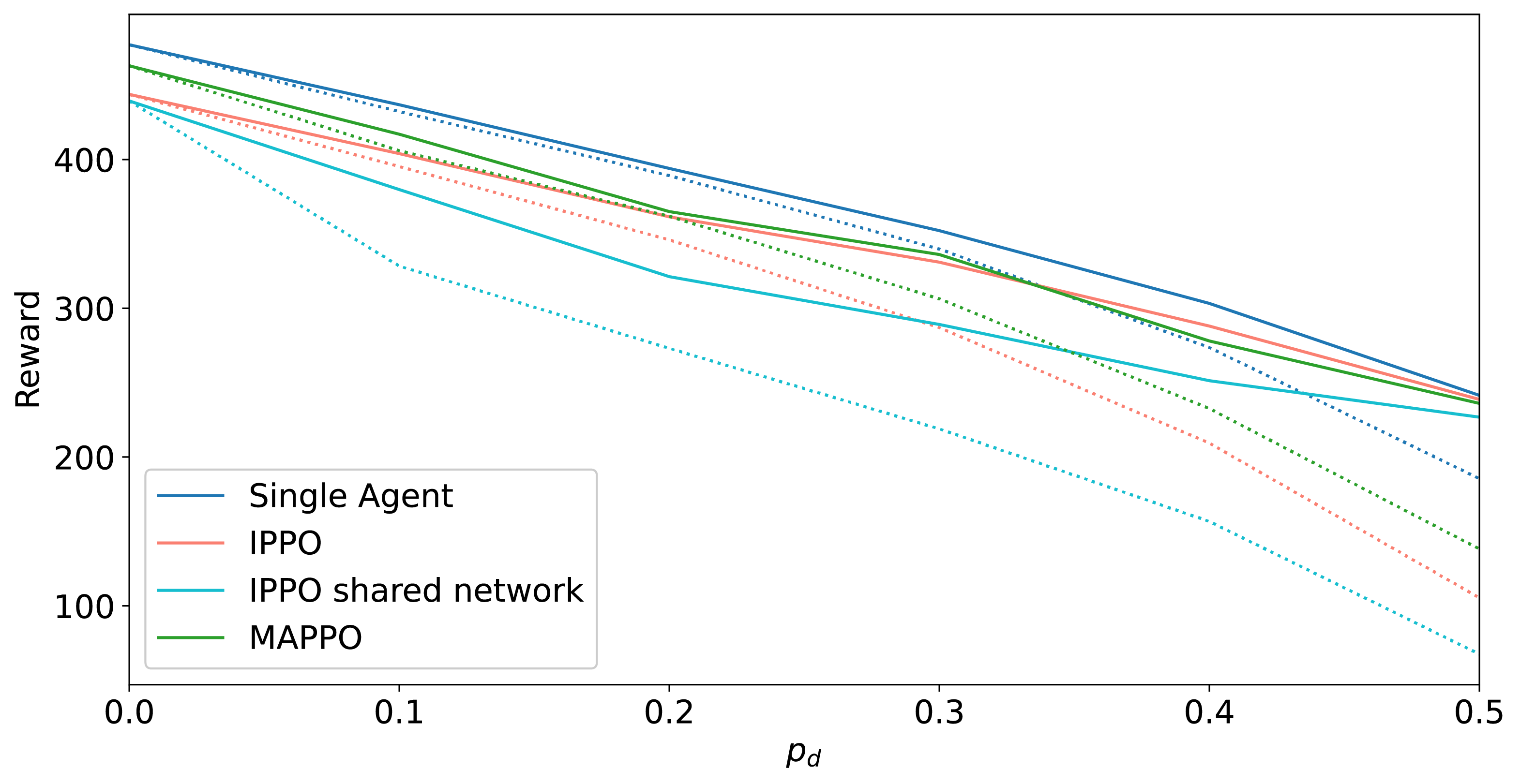}
\caption{Customer demand disturbance.}
\label{fig:demand noise trained}
\end{subfigure}%
\begin{subfigure}{.49\textwidth}
\centering
\includegraphics[width=0.99\linewidth]{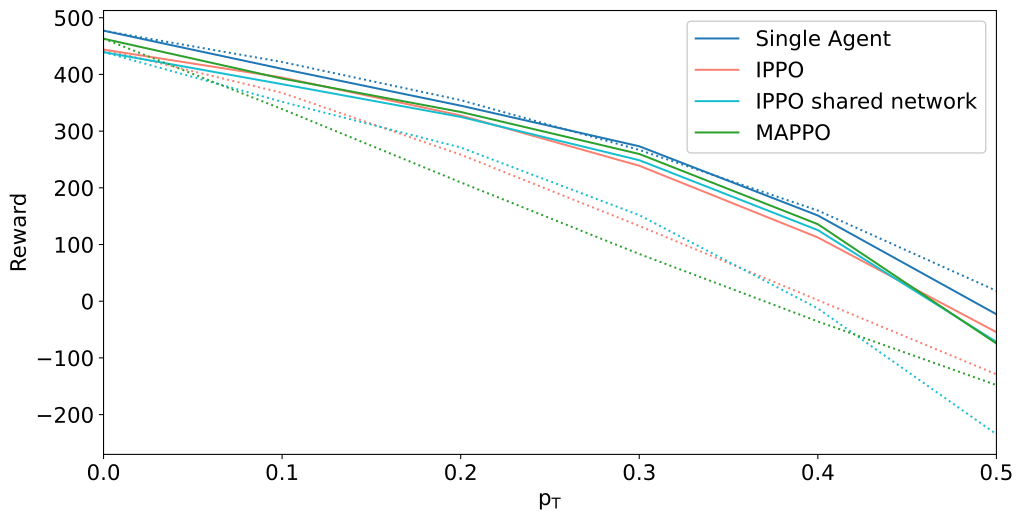}
\caption{Lead-time disturbance.}
\label{fig:delay noise trained}
\end{subfigure}
\caption{The performance difference of the multi-agent RL algorithms when training with \textbf{S2} against training with \textbf{S1} (dotted lines).}
\label{fig: noise trained}
\end{figure}

In this section, we investigate how uncertainty in the form of customer demand and lead time disturbances, as described above in \S\ref{sec: uncertainty}, affect the performance of the MARL methods. Previously, we investigated the performance of MARL methods when trained and evaluated on different uncertainty settings, \textbf{S2} and \textbf{S3}, and compared the results to the base-case four-stage serial SCN with \textbf{S1}. In this section, we train the MARL methods on the base case uncertainty settings \textbf{S1}, but then evaluate the policies under the uncertainty settings \textbf{S2} and \textbf{S3}. We compare their results to the centralized RL agent to assess the robustness of the distributed data-driven methods. This is shown in Figure \ref{fig:demand noise trained} for \textbf{S2} we plot the mean reward of the different algorithms at the different levels of disturbance, where the solid lines represent the algorithms trained on data with \textbf{S2} while the dotted lines represent the algorithms trained with \textbf{S1}.

We can see that the performance of all algorithms improves significantly when they are trained on data with non-stationary customer demand as opposed to when they are not, with this improvement being more significant at higher levels of $p_d$. When the algorithms are not trained on non-stationary customer demand, we can see that MARL methods are less robust to disturbances in customer demand than the centralized single agent, experiencing larger drops in performance. This could be due to the nature of the disturbance, which appears at the customer-facing node, paired with the limited observability of agents during execution. The centralized agent can observe the whole SCN, however for the MARL agents without communication, information flows between nodes at the speed of order flows, thus nodes further upstream cannot account for the disturbance before it is too late.  While using MAPPO or IPPO leads to a marginally larger drop in performance than the single agent, both algorithms perform worse in absolute terms. However, MARL algorithms experience a much larger performance gain than the single agent when trained on data with non-stationary customer demand and approach the performance of the single agent in absolute terms at higher noise levels. \par
The same behavior can be observed when using stochastic delivery lead times as shown in Figure \ref{fig:delay noise trained}. In this case, the effect is even more significant with MARL algorithms performing significantly worse when trained with deterministic lead times with the \textbf{S1} uncertainty setting. This is due to the fact that the state configurations used for MARL algorithms include incoming shipment information which becomes a forecast under lead-time uncertainty. However, training on data with stochastic delivery lead times in \textbf{S3} results in the MARL agents almost matching the performance of the centralized agent across a range of noise levels. This shows the robustness of MARL algorithms in terms of training (while utilizing the same algorithm configuration) which is particularly useful for applications such as SCNs where structural changes to the data-generating process happen over time and algorithms need to be re-trained.\par

\subsection{Independent Rewards}
\label{sec: independent rewards}
\begin{figure}
\centering
\begin{subfigure}{.49\textwidth}
\centering
\includegraphics[width=0.99\linewidth]{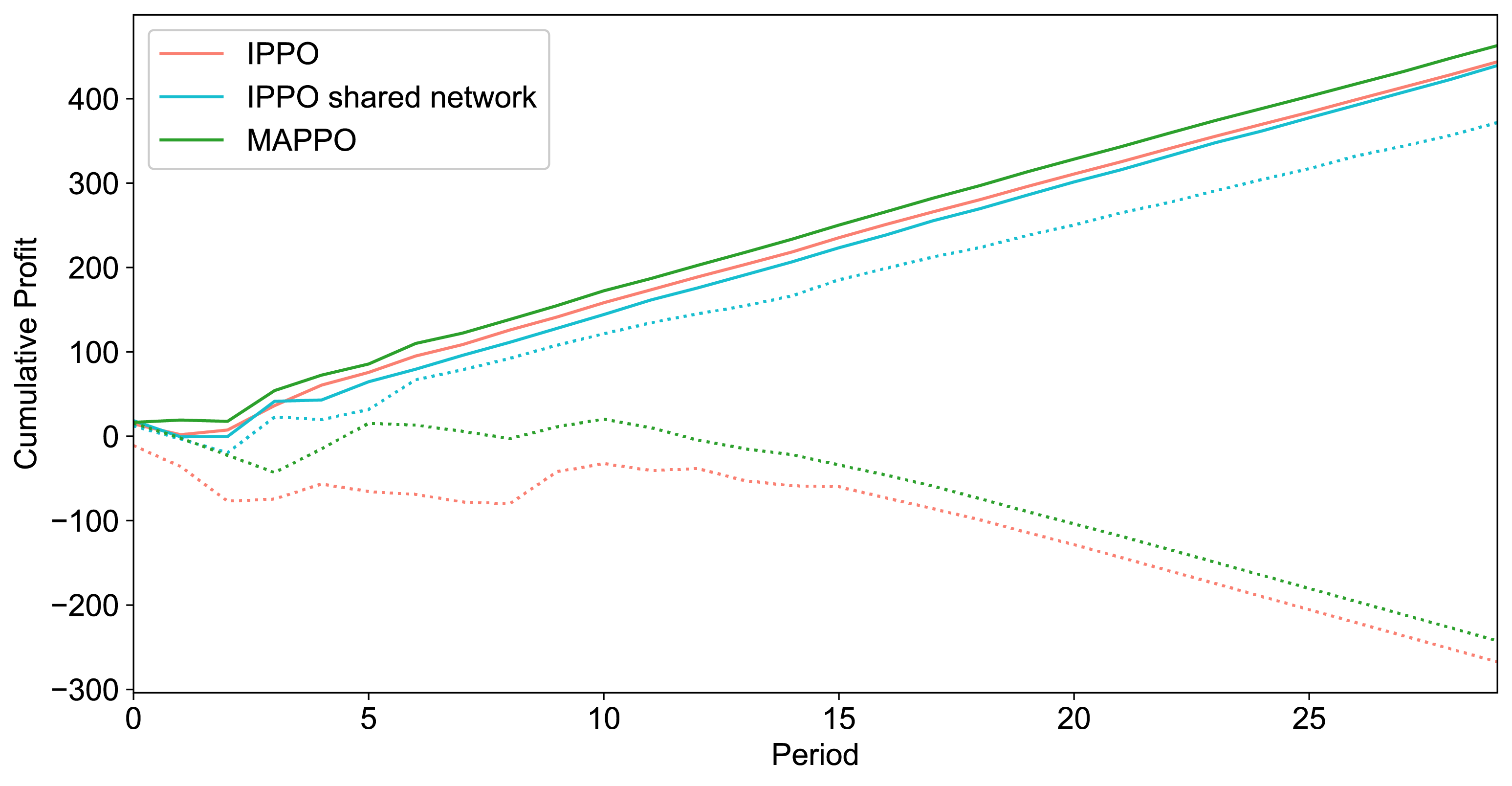}
\caption{Serial SCN.}
\label{fig: 4 stage independent rewards}
\end{subfigure}%
\begin{subfigure}{.49\textwidth}
\centering
\includegraphics[width=0.99\linewidth]{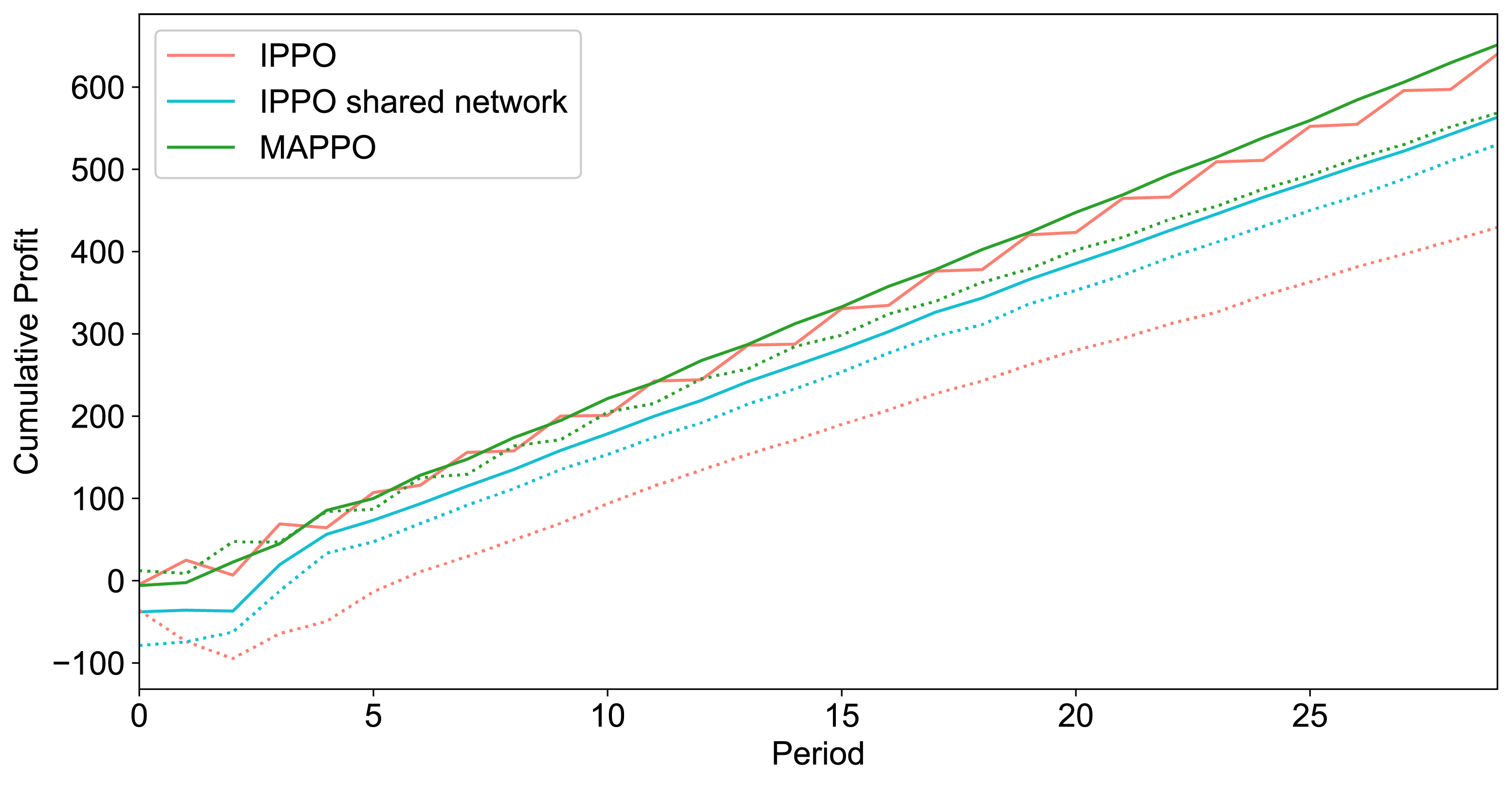}
\caption{Divergent SCN.}
\label{fig: div independent rewards}
\end{subfigure}
\caption{The effect of using independent rewards (dotted lines) on the multi-agent RL algorithms compared with the use of shared rewards (solid lines).}
\label{fig: independent rewards}
\end{figure}

In all the above simulations and analyses, MARL agents were set to maximize the total SCN profit as each agent receives the average total profit as a reward. Using shared rewards leads to more stable training as well as more coordination (especially in the absence of communication) as agents learn a policy aimed at maximizing the total SCN profit which would be desired as a real-life objective. However, in most real SCNs nodes will seek to maximize their own individual reward as each node is an independent entity with its own objectives. It was found, however, that training MARL agents with independent rewards lead to unstable training as well as significantly worse performance. We compare the performance (in terms of total SCN profit achieved) of the different MARL algorithms trained with shared rewards against those trained with independent rewards on 200 episodes to assess the effect of using independent rewards. This is shown in Figures \ref{fig: 4 stage independent rewards} and \ref{fig: div independent rewards} for the four-stage and divergent SCNs respectively where dotted lines represent the algorithms trained with independent rewards. 

It can be seen that in both cases, using IPPO with a shared network resulted in the smallest drop in performance where the agents were still able to learn a good policy. IPPO on the other hand experienced the largest drop in performance in both cases. Using a shared network might result in more coordination as the same network is used to find a reward maximizing policy for all agents even when utilizing independent rewards, whereas IPPO with independent rewards is completely decentralized. MAPPO also experiences a large drop in performance, especially in the four-stage serial SCN, even while not being completely decentralized due to the use of global observation for the critic network. That being said, the MAPPO algorithm was still able to learn a decent policy for the divergent SCN with independent rewards.

\subsection{Summary}
Three different MARL methods were analyzed in different SCN configurations to assess their effectiveness for the IM problem. During online execution, MARL methods are orders of magnitude faster online as compared with traditional model-based LPs. While this faster execution comes at the cost of offline training, it also provides complete decentralization with no requirement to coordinate computation online. The computational efficiency of trained agents also becomes more important as the size of SCNs, and hence the IM problem, becomes larger and more complex. In terms of performance, all the MARL methods considered outperform a corresponding distributed LP method to the problem in serial SCNs as seen in \S\ref{sec: base case}. While this improvement in performance did not extend to divergent SCNs as seen in \S\ref{sec: divergent chain}, IPPO and MAPPO still learn a good policy while having the added advantage of being orders faster during online execution than DSHLP. In terms of robustness to external uncertainty in \S\ref{sec: uncertainty}, we find that due to their data-driven model-free nature, the MARL methods were much more robust to lead time uncertainty than DSHLP and manage to outperform it over a range of customer demand disturbance levels. 
Of the MARL methods tested, MAPPO and IPPO were found to have consistently good performance, almost matching that of an equivalent centralized implementation. Using IPPO with a shared network, however, leads to performance drops for larger SCNs such as the eight-stage serial SCN, or SCNs where all nodes don't share similar dynamics like the divergent SCN in \S\ref{sec: divergent chain}. When using independent rewards as in \S\ref{sec: independent rewards} it was found that using IPPO with a shared network still leads to a good policy in terms of total SCN profit with a significantly smaller drop in performance than MAPPO and IPPO.

\section{Conclusions}
Three distributed data-driven methods utilizing MARL were proposed (IPPO, IPPO with shared network and MAPPO) to find a dynamic re-order policy for IM problem. This offers a more practical solution to real-world supply-chain networks as they are made up of independent entities. While it was hypothesized that the use of RNNs would improve the performance of MARL agents given their partial observability, it was found that using RNNs did not lead to better policies. The MARL methods were assessed across different SCN configurations and the results have shown that they achieve a performance close to that of the single RL agent centralized solution in terms of total profit achieved. Furthermore, they are more scalable as the solution space of the agents in MAPPO and IPPO does not grow with the size of the problem, with MAPPO outperforming the centralized RL for an eight-stage serial SCN. They also outperform the distributed solution utilizing LP for most cases tested and are more robust to demand and lead-time uncertainty while being orders more computationally efficient during execution. We show that having full observation during training leads to better performance where MAPPO achieved the highest mean rewards across the different SCN configurations tested. IPPO performs similarly to MAPPO but we find that sharing a policy between agents leads to worse performance in the case with IPPO with a shared network.\par
The results of the proposed MARL solution to the IM problem show decentralized data-driven control provides a good control solution to large-scale stochastic systems. They perform nearly as well as their centralized counterpart while making fewer assumptions and respecting real-world information-sharing constraints. This makes them far more practical and applicable to real SCNs. Furthermore, this shows that MARL is a viable solution for many other similar problems in OR.

In order for the MARL method to handle the more complex systems some improvements would need to be made. The agents' handling of system disruptions could be improved through the use of differentiable communication channels \citep{CommNet} which will allow agents to learn to communicate useful information. This would reduce the problem caused by the partial observability of agents as information can be propagated through the system faster than the flow of orders and materials. The use of Graph Neural Networks (GNN) in particular could make use of the underlying SCN graph structure to allow for more efficient and effective communication. GNNs have been shown to improve performance for collaborative agents \citep{Jiang2020Graph, HeteroGNNRL} as it allows for better coordination. Furthermore, the use of vertical federated RL \cite{qi2021federated} could help overcome the problem of centralization of information during training.

\section*{Acknowledgements}
The authors would like to acknowledge the contribution and feedback of Dr. Dongda Zhang, which helped to improve the clarity of the paper. Max Mowbray acknowledges support from the Engineering and Physical Sciences Research Council grant EP/T517823/1.

 \bibliographystyle{elsarticle-harv} 
 \bibliography{cas-refs}






\appendix
\section{Algorithm Description}
\label{sec: Algorithm}
In this section, we describe the PPO algorithm utilized in this work. Like regular policy gradient methods, the aim is to find a policy $\pi(a|s)$ parameterized by a neural network with parameters $\theta$ that maximizes total episode returns. A surrogate objective $L$ is maximized such that the policy update is given by
\begin{equation}
    \theta_{k+1} = \arg\max_{\theta}\mathbb{E}_{s,a \sim \pi_{\theta_{k}}}[L(s, a, \theta_k, \theta)],
\end{equation}
The aim of PPO is to avoid taking steps in parameter space $\theta$ that may cause the policy $\pi_{\theta}(a|s)$ to collapse by ensuring that the new policy does not deviate too much from the old policy. \par
Therefore, the PPO-clip algorithm utilizes the following surrogate objective.

\begin{equation}
\resizebox{0.45\textwidth}{!}{$
    L^{\text{clip}}(s, a, \theta_k, \theta) = \min \Bigg(
    \frac{\pi_{\theta}(a|s)}{\pi_{\theta_{k}}(a|s)}A^{\pi_{\theta_{k}}}, \hspace{2mm}
    \text{clip}\Bigg(
    \frac{\pi_{\theta}(a|s)}{\pi_{\theta_{k}}(a|s)}, 1-\epsilon, 1+\epsilon
    \Bigg)
    A^{\pi_{\theta_{k}}}
    \Bigg),
    $}
\end{equation}

where $A^{\pi_{\theta_{k}}}$ is the advantage function under policy $\pi_{\theta_{k}}$ which describes the advantage of taking action $a$ in state $s$ according to $\pi(\cdot|s)$ over the policy's average action in that state.
The PPO-clip surrogate objective forces the ratio between the old and new policy to stay with an interval $[1-\epsilon, \hspace{1mm} 1+\epsilon]$, where the policy clip parameter $\epsilon$ is a hyperparameter, through the use of the \emph{clip} function.\par
In addition to the PPO-clip objective, an adaptive KL penalty coefficient was used which adds a penalty on the KL divergence, $d_{KL}$ between the old and new policies and adapts the penalty coefficient until a target KL divergence, $d_{\text{targ}}$ is achieved. The adaptive KL penalty objective is given by \citep{PPO} as
\begin{equation}
    L^{\text{KL}}(s, a, \theta_k, \theta) = \frac{\pi_{\theta}(a|s)}{\pi_{\theta_{k}}(a|s)}A^{\pi_{\theta_{k}}} - \beta \text{KL}[\pi_{\theta_{k}}(\cdot|s), \pi_{\theta}(\cdot|s)],
\end{equation}
where $\beta$ is the adaptive KL divergence coefficient. At each policy improvement step, the KL divergence between the old and new policies is calculated. If $d_{KL}<\frac{2}{3}d_{\text{targ}}$ then $\beta \leftarrow \frac{\beta}{2}$ and if  $d_{KL}>\frac{3}{2}d_{\text{targ}}$ then $\beta \leftarrow \beta \times 2$. Therefore, the surrogate objective utilized in the PPO implementation for this work is given by
\begin{equation}
    L(s, a, \theta_k, \theta) = L^{\text{clip}} - \beta \text{KL}[\pi_{\theta_{k}}(\cdot|s), \pi_{\theta}(\cdot|s)]
    \label{eqn: ppo objective}
\end{equation}
A good estimate of $A^{\pi}$ reduces the variance of the policy gradient and thus results in fewer samples being required. In this work, the \emph{generalized advantage estimator} (GAE) \citep{gae} was used to estimate the advantage function which is given by
\begin{equation}
    A^{\pi}_{\text{GAE}}(s_t, a_t) = \sum^{\infty}_{l=0}(\gamma \lambda)^l \delta_{t+l}
    \label{eqn: gae}
\end{equation}
where
\begin{equation*}
    \delta_t = r_t + \gamma V^{\pi}(s_{t+1}) - V^{\pi}(s_{t})
\end{equation*}
\\
\noindent
is the TD-error or advantage function for time-step $t$ with 1-step forward returns, $\gamma$ is the discount factor and $\lambda$ is a hyperparameter that controls the bias-variance trade-off of the GAE estimate. Equation (\ref{eqn: gae}) is, in essence, a weighted average of several advantage estimators with different biases and variances. The 1-step advantage estimate $\delta_t$ is high-bias and low-variance while higher $l$-step advantages have lower bias but higher variance. Contributions from $l$-step advantages decay exponentially where the decay rate is controlled by $\lambda$. A higher value of $\lambda$, therefore, results in higher variance and lower bias and vice-versa.\par
The GAE requires values for the value function $V$ which is estimated using a separate neural network with parameters $\phi$. This leads to an actor-critic structure where the actor-network $\pi$ updates the policy $\pi_{\theta}$ using the objective function (\ref{eqn: ppo objective}) while the critic estimates the value function of the system's states $V_{\phi}$ in order to provide more accurate advantage estimates for policy optimization. The algorithm used in this work is summarised below. \par

\begin{algorithm}[H]
\caption{PPO clip with adaptive KL penalty \citep{SpinningUp2018}}\label{alg:ppo}.
\begin{algorithmic}[1]
\State Input: Initial policy parameters $\theta_0$ and initial value function parameters $\phi_0$
\For{$k=0,1,2, \dots$} 
\State Collect set of trajectories $\mathcal{D}$ = $\{s_i, a_i, r_i, s'_i \}$ by running policy $\pi_{\theta_{k}}$ in the environment.
\State Compute rewards-to-go $R_t$.
\State Compute GAE advantage estimates $\hat{A}_t$ using the current value function estimates $V_{\phi_{k}}$ over several mini-batches using stochastic gradient ascent with Adam. \citep{kingma2017adam}.
\State Fit value function using mean-square error
\begin{equation*}
    \phi_{k+1} = \arg\min_{\phi}\frac{1}{|\mathcal{D}_k|T}\sum_{\tau \in \mathcal{D}_k} \sum_{t=0}^T \big(V_{\phi}(s_t) - R_t \big)^2
\end{equation*}
\hspace{5.8mm}over several mini-batches using stochastic gradient descent with Adam.
\State Update the policy by maximizing the surrogate objective:
\begin{equation*}
    \theta_{k+1} = \arg\max_{\theta}\frac{1}{|\mathcal{D}_k|T}\sum_{\tau \in \mathcal{D}_k} \sum_{t=0}^T \big(L^{\text{clip}} - \beta \text{KL}[\pi_{\theta_{k}}(\cdot|s), \pi_{\theta}(\cdot|s)]\big)
\end{equation*}

\EndFor
\end{algorithmic}
\end{algorithm}

\section{Agent Training}
\label{sec: Agent Training}
In order to determine which variables to include in the state vector or which neural network architecture from Figure \ref{fig: Single agent Network architecture} we train RL agents for each of the MARL algorithms being investigated on different combinations and compare their performance. To allow for a fair comparison, we tune the hyperparameters of the algorithms for each combination and test their performance for one SCN environment. The hyperparameter tuning is described in more detail in \ref{sec: Hyperparameter Tuning} and the final values used for the algorithms are shown in \ref{sec: Hyperparameter Values}.

A four-stage serial SCN was chosen as the base environment over which to tune the hyperparameters with the environment parameters shown below in Table \ref{tab:four stage config} with an episode containing a number of periods $T=30$. Customer demand was assumed to follow a Poisson distribution with $\lambda_{\text{Poisson}}=5$.

\begin{table}[H]
  \centering
  \caption{IM environment configuration for a four-stage supply chain.}
  \resizebox{0.7\textwidth}{!}{
    \begin{tabular}{lccccc}
    \hline
    \textbf{Parameter} & \textbf{Symbol} & \textbf{Node 1} & \textbf{Node 2} & \textbf{Node 3} & \textbf{Node 4} \\
    \hline
    Initial inventory & $i[0]$  & 10    & 10    & 10    & 10 \\
    Sell price & $P$     & 2     & 3     & 4     & 5 \\
    Replenishment cost & $C$     & 1     & 2     &3     &4 \\
    Storage cost & $I$     & 0.35  & 0.30   & 0.40   & 0.20 \\
    Backlog cost & $B$     & 0.50   & 0.70   & 0.60   & 0.90 \\
    Storage capacity & $I_{\max}$     & 30    & 30    & 30    & 30 \\
    Order limit & $O_{r_{\max}}$     & 30    & 30    & 30    & 30 \\
    Lead time & $\tau$   & 1     & 2     & 3     & 1 \\
    \end{tabular}
    }%
  \label{tab:four stage config}%
\end{table}%

In order to decide on which variables to include in the state vector, different combinations of the state vector groups shown in Figure \ref{fig: state} were tested. The different state vector configurations tested are:
\begin{enumerate}
    \item Base State
    \item Base State + Delayed Shipment
    \item Base State + Previous Demand + Previous Orders
    \item Base State + Previous Demand
    \item Base State + Previous Demand + Previous Orders +  Delayed Shipment
    \item Base State + Previous Demand +  Delayed Shipment
\end{enumerate}

After training each agent state vector configuration from the above with both neural network architectures, to completion, each trained set of agents was tested on the same 1000 test episodes each consisting of a realization of the customer demand over 30 time steps. The mean episode reward for each configuration, along with its standard deviation is shown below for IPPO in Table \ref{tab:IPPO config results} and for MAPPO in Table \ref{tab:MAPPO config results}.

\begin{table}[H]
  \centering
  \caption{IPPO performance of each state-vector configuration/network architecture combination.}
    \resizebox{0.6\textwidth}{!}{
    \begin{tabular}{lp{2.5em}p{2.5em}p{2.5em}p{2.5em}p{2.5em}p{2.5em}}
    \hline
    \textbf{State Configuration} & \multicolumn{1}{c}{\textbf{1}} & \multicolumn{1}{c}{\textbf{2}} & \multicolumn{1}{c}{\textbf{3}} & \multicolumn{1}{c}{\textbf{4}} & \multicolumn{1}{c}{\textbf{5}} & \multicolumn{1}{c}{\textbf{6}} \\
    \hline
    \textbf{Without RNN} & 410.6\newline{}(68.7) & 439.9\newline{}(81.1) & 409.6\newline{}(62.7) & 419.5\newline{}(61.7) & 442.9\newline{}(65.5) & \underline{450.3}\newline{}(66.6) \\
    & & & & & &\\
    \textbf{With RNN} & 422.2\newline{}(74.3) & 413.6\newline{}(81.2) & 414.8\newline{}(84.3) & 429.1\newline{}(70.5) & 434.6\newline{}(78.9) & 397.1\newline{}(93.4) \\
    \end{tabular}
    }%
  \label{tab:IPPO config results}%
\end{table}%

\begin{table}[H]
  \centering
  \caption{MAPPO performance of each state-vector configuration/network architecture combination.}
    \resizebox{0.6\textwidth}{!}{
    \begin{tabular}{lp{2.5em}p{2.5em}p{2.5em}p{2.5em}p{2.5em}p{2.5em}}
    \hline
    \textbf{State Configuration} & \multicolumn{1}{c}{\textbf{1}} & \multicolumn{1}{c}{\textbf{2}} & \multicolumn{1}{c}{\textbf{3}} & \multicolumn{1}{c}{\textbf{4}} & \multicolumn{1}{c}{\textbf{5}} & \multicolumn{1}{c}{\textbf{6}} \\
    \hline
    \textbf{Without RNN} & 398.7\newline{}(62.6) & 407.0\newline{}(81.0) & 423.8\newline{}(61.4) & 400.3\newline{}(61.1) & \underline{436.7}\newline{}(60.5) & 430.3\newline{}(72.4) \\
    & & & & & &\\
    \textbf{With RNN} & 440.0\newline{}(89.8) & 414.7\newline{}(91.3) & 400.4\newline{}(88.5) & 424.0\newline{}(73.5) & 396.5\newline{}(76.5) & 425.0\newline{}(75.9) \\
    \end{tabular}
    }%
  \label{tab:MAPPO config results}%
\end{table}%

While it was hypothesised that the use of an RNN based network would significantly improve the performance of the multi-agent system due to each agent's limited observability, the best performing combination for the IPPO (underlined in Table \ref{tab:IPPO config results} above) did not utilise an RNN. It can also be seen that state vector configurations that included more information through the use of more variables seemed to perform better without the use of an RNN. One explanation could be that knowledge of variables such as previous demand and actions is sufficient to capture enough of the system's history such that actions can be conditioned on just the observation without the need for the RNN's internal hidden state. 

For MAPPO, the best performing state vector configuration and network architecture combination was actually configuration 1 with the use of an RNN. However, it was only marginally better than using state vector configuration 5 without an RNN. It also had a significantly higher standard deviation of rewards which lead to using state vector configuration 5 without an RNN instead. \par

The same analysis and training were done for a centralized RL agent controlling the entire SCN, utilizing the same PPO algorithm described in \ref{sec: Algorithm}. The results are shown in Table \ref{tab:single agent config results} below. It can be seen that state vector configuration 3 with a neural network not utilizing an RNN achieved the best episode mean reward. This setting is used by the centralized single RL agent in all the results shown in this work.

\begin{table}[htbp]
  \centering
  \caption{Single-agent performance of each state-vector configuration/network architecture combination.}
    \resizebox{0.7\textwidth}{!}{
    \begin{tabular}{lp{2.5em}p{2.5em}p{2.5em}p{2.5em}p{2.5em}p{2.5em}}
    \hline
    \textbf{Network Architecture} & \multicolumn{1}{c}{\textbf{1}} & \multicolumn{1}{c}{\textbf{2}} & \multicolumn{1}{c}{\textbf{3}} & \multicolumn{1}{c}{\textbf{4}} & \multicolumn{1}{c}{\textbf{5}} & \multicolumn{1}{c}{\textbf{6}} \\
    \hline
    \textbf{Without RNN} & 465.1\newline{}(60.4) & 476.3\newline{}(53.9) & \underline{478.2}\newline{}(47.7) & 437.6\newline{}(34.7) & 453.6\newline{}(43.1) & 465.8\newline{}(35.4) \\
    & & & & & &\\
    \textbf{With RNN} & 438.2\newline{}(58.5) & 453.1\newline{}(57.8) & 434.6\newline{}(73.2) & 453.6\newline{}(59.7) & 455.6\newline{}(55.9) & 446.0\newline{}(58.4) \\
    \end{tabular}
    }%
  \label{tab:single agent config results}%
\end{table}%

\section{Hyperparameter Optimization}
\label{sec: Hyperparameter Tuning}
 The hyperparameters tuned include PPO-specific hyperparameters such as the number of experiences/time-step collected in every iteration i.e. the batch size $|\mathcal{D}|$ and subsequently the number of epochs and minibatch size for the SGD as well the learning rate for the Adam optimizer. Other hyperparameters include variables relating to the network architecture, particularly the number of neurons in each hidden layer. This is summarised below along with the range of values considered for each hyperparameter in Table \ref{tab: hyperam settings}. For a network without an RNN, the hyperparameter search setting is identical to the one shown below but without the RNN-related hyperparameters (Actor/Critic Encoding layer size and LSTM state size). It is worth noting that ranges in parenthesis refer to search space over a uniform or uniform integer distribution with the parameters of the distribution in the parentheses, whereas ranges in square brackets involve searching among the distinct values within the brackets.\par

To find the best hyperparameters, the Ray \emph{tune} library \cite{raytune} was used. Within the Ray tune framework, a \emph{population-based training} (PBT) \cite{ppt} scheduler was used to optimize the hyperparameter values. PBT involves training many neural networks or, in this case, agents, in parallel starting off with random hyperparameter values from prescribed ranges. It then goes for different phases of exploration or exploitation of hyperparameters. In exploration, a new value for a hyperparameter is sampled from the given distribution, whereas during exploitation the parameters of a better-performing agent might be copied.\par

In this work, 6 agents were trained in parallel. Hyperparameters were perturbed every 4 training iterations with workers training each agent deciding on whether to explore or exploit. This process was stopped after 200 training iterations as preliminary experiments had shown this is sufficient for policy convergence.

\begin{table}[H]
  \centering
  \caption{Multi-agent centralized critic hyperparameter optimization settings.}
    \begin{tabular}{lc}
    \hline
    \textbf{Hyperparameter} & \textbf{Value range} \\
    \hline
    Batch size, $|\mathcal{D}|$ & (1500, 9000) \\
    Minibatch size & (64, 256) \\
    Epochs & (3, 30) \\
    Clip parameter, $\epsilon$ & (0.1, 0.4) \\
    Discount factor, $\gamma$& (0.95, 0.99) \\
    GAE parameter, $\lambda$& (0.90, 1.0) \\
    Initial KL coefficient, $\beta$ & (0.1, 0.6) \\
    KL target, $d_{\text{targ}}$ & (0.003, 0.03) \\
    Learning rate, $\alpha$ & [5e-4, 1e-4 5e-5, 1e-5, 5e-6] \\
    FC1 size & [64, 128, 256] \\
    FC2 size & [64, 128, 256] \\
    Actor Encoding size & [16, 32, 64, 128, 256] \\
    Critic Encoding size & [16, 32, 64, 128, 256] \\
    LSTM state size & [64, 128, 256] \\
    Previous length, $N$ & [1, 2, 3] \\
    \end{tabular}%
  \label{tab: hyperam settings}%
\end{table}%

\section{Hyperparameter Values}
\label{sec: Hyperparameter Values}
The hyperparameter values for each of the RL algorithms used are shown below in Tables \ref{tab:single agent final hyperparams}, \ref{tab:MAPPO agent final hyperparams} and \ref{tab:IPPO agent final hyperparams}. It is worth noting that the IPPO hyperparameter values were used for both the IPPO and IPPO with shared network algorithms.
\begin{table}[H]
  \centering
  \caption{IPPO hyperparameter values.}
  \resizebox{0.5\textwidth}{!}{
    \begin{tabular}{lc}
    \hline
    \textbf{Hyperparameter} & \textbf{Value range} \\
    \hline
    Batch size, $|\mathcal{D}|$ & 3000 \\
    Minibatch size & 180 \\
    Epochs & 14 \\
    Clip parameter, $\epsilon$ & 0.21 \\
    Discount factor, $\gamma$& 0.99 \\
    GAE parameter, $\lambda$& 0.95 \\
    Initial KL coefficient, $\beta$ & 0.19 \\
    KL target, $d_{\text{targ}}$ & 0.006 \\
    Learning rate, $\alpha$ & $1 \times 10^{-5}$ \\
    FC1 size & 128 \\
    FC2 size & 128 \\
    \end{tabular}
    }%
  \label{tab:IPPO agent final hyperparams}%
\end{table}%
\begin{table}[H]
  \centering
  \caption{MAPPO hyperparameter values.}
  \resizebox{0.5\textwidth}{!}{
    \begin{tabular}{lc}
    \hline
    \textbf{Hyperparameter} & \textbf{Value range} \\
    \hline
    Batch size, $|\mathcal{D}|$ & 4147 \\
    Minibatch size & 92 \\
    Epochs & 12 \\
    Clip parameter, $\epsilon$ & 0.41 \\
    Discount factor, $\gamma$&  0.966 \\
    GAE parameter, $\lambda$& 0.973 \\
    Initial KL coefficient, $\beta$ & 0.69 \\
    KL target, $d_{\text{targ}}$ & 0.003 \\
    Learning rate, $\alpha$ & $1 \times 10^{-5}$ \\
    FC1 size & 256 \\
    FC2 size & 256 \\
    \end{tabular}
    }%
  \label{tab:MAPPO agent final hyperparams}%
\end{table}%
\begin{table}[H]
  \centering
  \caption{Centralized single RL agent hyperparameter values.}
  \resizebox{0.5\textwidth}{!}{
    \begin{tabular}{lc}
    \hline
    \textbf{Hyperparameter} & \textbf{Value range} \\
    \hline
    Batch size, $|\mathcal{D}|$ & 3000 \\
    Minibatch size & 225 \\
    Epochs & 10 \\
    Clip parameter, $\epsilon$ & 0.22 \\
    Discount factor, $\gamma$& 0.99 \\
    GAE parameter, $\lambda$& 0.95 \\
    Initial KL coefficient, $\beta$ & 0.37 \\
    KL target, $d_{\text{targ}}$ & 0.0043 \\
    Learning rate, $\alpha$ & $1 \times 10^{-5}$ \\
    FC1 size & 64 \\
    FC2 size & 256 \\
    Previous length, $N$ & 1 \\
    \end{tabular}
    }%
  \label{tab:single agent final hyperparams}%
\end{table}%

\section{Supply Chain Network Environment Configurations}
\label{sec: Environment Config}
The tables below summarise the environment parameters for the different SCN configurations used.

\begin{table}[H]
  \centering
  \caption{The IM environment configuration of the divergent supply chain.}
  \resizebox{0.7\textwidth}{!}{
    \begin{tabular}{lccccc}
    \hline
    \textbf{Parameter} & \textbf{Symbol} & \textbf{Node 1} & \textbf{Node 2} & \textbf{Node 3} & \textbf{Node 4} \\
    \hline
    Initial inventory & $i[0]$  & 10    & 10    & 10    & 10 \\
    Sell price & $P$     & 2     & 3     & 4     & 4 \\
    Replenishment cost & $C$     & 1     & 2     &3     &3 \\
    Storage cost & $I$     & 0.35  & 0.30   & 0.40   & 0.40 \\
    Backlog cost & $B$     & 0.50   & 0.70   & 0.60   & 0.60 \\
    Storage capacity & $I_{\max}$     & 30    & 30    & 30    & 30 \\
    Order limit & $O_{r_{\max}}$     & 30    & 30    & 30    & 30 \\
    Lead time & $\tau$   & 1     & 2     & 1     & 1 \\
    \end{tabular}
    }%
  \label{tab:div config}%
\end{table}%

\begin{table}[H]
  \centering
  \caption{IM environment configuration for a two-stage supply chain.}
  \resizebox{0.4\textwidth}{!}{
    \begin{tabular}{ccc}
    \hline
    \textbf{Symbol} & \textbf{Stage 1} & \textbf{Stage 2} \\
    \hline
    $i[0]$  & 10    & 10 \\
    $P$ & 3     & 2 \\
    $C$ & 2     & 1 \\
    $I$ & 0.50  & 0.20  \\
    $B$ & 0.60   & 0.90 \\
    $I_{\max}$ & 30    & 30 \\
    $O_{r_{\max}}$ & 30    & 30 \\
    $\tau$  & 3  & 1 \\
    \end{tabular}
    }%
  \label{tab:two stage config}%
\end{table}%

\begin{table}[H]
  \centering
  \caption{IM environment configuration for an eight-stage supply chain.}
  \resizebox{0.75\textwidth}{!}{
    \begin{tabular}{ccccccccc}
    \hline
    \textbf{Symbol} & \textbf{Stage 1} & \textbf{Stage 2} & \textbf{Stage 3} & \textbf{Stage 4} & \textbf{Stage 5} & \textbf{Stage 6} & \textbf{Stage 7} & \textbf{Stage 8}\\
    \hline
    $i[0]$ &10 &10 &10 &10 &10 &10 &10 &10 \\
    $P$ &9 &8 &7 &6 &5 &4 &3 &2 \\
    $C$ &8 &7 &6 &5 &4 &3 &2 &1 \\
    $I$ &0.35 &0.30 &0.40 &0.20 &0.35 &0.30 &0.40 &0.20 \\
    $B$ &0.50 &0.70 &0.60 &0.90 &0.50 &0.70 &0.60 &0.90 \\
    $I_{\max}$ &30 &30 &30 &30 &30 &30 &30 &30 \\
    $O_{r_{\max}}$ &30 &30 &30 &30 &30 &30 &30 &30 \\
    $\tau$   &1 &2 &3 &1 &4 &2 &3 &1\\
    \end{tabular}
    }%
  \label{tab:eight stage config}%
\end{table}%

\section{Mathematical Programming Benchmarks}
\label{sec: OR methods}
Linear Programming (LP) was used to find solutions for the IM problem to compare the performance of traditional model-based re-order policies against that of the data-driven model-free RL methods developed. Dynamic re-order policies involve finding the optimal replenishment order level at every time step by solving the optimization problem above. In this work, the multi-period IM problem was solved using a shrinking horizon where at every time-step $j$, the objective to be maximized is
\begin{equation}
    \max \sum_{t=j}^T \sum_{m=1}^M  P^m s_r^m[t] - C^m o_r^m[t] - I^m i^m[t] - B^m b^m[t].
\end{equation}

The only difference to the original optimization problem is that the demand for customer-facing nodes is now equal to the expectation of customer demand $\forall t>j$ for future time steps. The first replenishment order for each node $o_r^m[j]$ from the solution to the above optimization problem is enacted while the actions for the remaining time periods are discarded. Once a step is taken and real customer demand is realized, the problem is re-initialized and solved in the next time period over a now shorter horizon as with MPC until $j=T$. \par
In order to compare this model-based optimization method against our MARL implementation, we used a distributed version of the above. By inspection of Equations (\ref{eqn: optimisation}) and (\ref{eqn: order demand equality}), we see that the problem becomes separable into M node-level sub-problems after relaxation of constraints (\ref{eqn: backlog}), (\ref{eqn: sales constraint 2}), (\ref{eqn: acquisition}) and (\ref{eqn: order demand equality}) where the problem is decomposed into node-level sub-problems that each agent can solve in private and parallel.\par
We use the \emph{Alternating Direction Method of Multipliers} (ADMM) under its consensus form for distributed optimization \citep{Boyd2010}: Rather than directly relax complicating constraints, we add the shared variables $z^{m_d}_1[t]$ and $z^{m_d}_2[t]$ as well as scaled dual variables $u^{m_d}_1[t]$ and $u^{m_d}_2[t]$ for each connection between nodes and each time step. The deviation between $z^{m_d}_1[t]$ and local copies $d_r^{m_d}[t]$ and $o_r^{m}[t]$ as well as between $z^{m_d}_2[t]$ and local copies $s_r^{m_d}[t]$ and $a_r^{m}[t]$ is then penalised in the objective of the $m^{th}$ sub-problem as follows:
\begin{equation}\small
\begin{split}
    \max & \sum_{t=j}^T \biggl( P^m s_r^m[t] - C^m o_r^m[t] - I^m i^m[t] - B^m b^m[t] - \frac{\rho}{2} 
    \\
    & \sum_{d \in \mathcal{D}_m}  \biggl[ 
    \bigl( o_r^{m}[t] - z_1^{m_d}[t] + u_1^{m_d}[t] \bigr)^2 + 
    \bigl( d_r^{m_d}[t] - z_1^{m_d}[t] + u_1^{m_d}[t] \bigr)^2 
    \\
    & + \bigl(a_r^{m}[t] - z_2^{m_d}[t] + u_2^{m_d}[t] \bigr)^2 + 
    \bigl( s_r^{m_d}[t-\tau^{m_d}] - z_2^{m_d}[t] + u_2^{m_d}[t] \bigr)^2 \biggr] \biggr)
    \label{eqn: optimization2}
\end{split}
\end{equation}

ADMM by consensus works as follows \citep{Boyd2010}:
At each iteration, the shared and scaled dual variables ($z^{m_d}_1[t]$, $z^{m_d}_2[t]$, $u^{m_d}_1[t]$, $u^{m_d}_2[t]$) are fixed and each node $m$ solves its own sub-problem consisting of objective (\ref{eqn: optimisation}) including only the relevant constraints (\ref{eqn: inventory})-(\ref{eqn: order demand equality}) to the current node. In these sub-problems, each node has total freedom over their copies of $d_r^{m_d}[t]$, $o_r^{m}[t]$, $s_r^{m_d}[t]$ and $a_r^{m}[t]$. Then, the shared variables ($z^{m_d}_1[t]$ and $z^{m_d}_2[t]$) are updated in a coordination step by averaging the optimal local copies for $d_r^{m_d}[t]$, $o_r^{m}[t]$, $s_r^{m_d}[t]$ and $a_r^{m}[t]$ between the two participating nodes of each connection. While the update in the shared variables aims to ensure asymptotic primal feasibility, the update in the dual variables aims to ensure asymptotic dual feasibility. Each agent's dual variables are updated based on the difference between the updated shared variables and each agent's optimal local copy thereof.

\end{document}